%% file: main.tex
\documentclass{article}

\PassOptionsToPackage{svgnames,dvipsnames}{xcolor}
\usepackage[final]{neurips_2025}

\input{math_commands.tex}

\usepackage[utf8]{inputenc} %
\usepackage[T1]{fontenc}    %
\usepackage{hyperref}       %
\usepackage{url}            %
\usepackage{booktabs}       %
\usepackage{amsmath}
\usepackage{amsthm}
\usepackage{amsfonts}       %
\usepackage{nicefrac}       %
\usepackage{microtype}      %
\usepackage{dsfont}
\usepackage{xcolor}         %

\newtheorem{definition}{Definition}

\newcommand{\norm}[1]{\left\lVert #1 \right\rVert}
\DeclareMathOperator{\sg}{sg} %
\newcommand{\I}{\mathbb{I}}
\DeclareMathOperator{\GeLU}{GeLU}

\newcommand{\codeLink}[0]{\href{https://github.com/harpoonix/GC-xLSTM}{github.com/harpoonix/GC-xLSTM}}

\newcommand{\denseColumns}[0]{\setlength{\tabcolsep}{5.1pt}}
\newcommand{\denserColumns}[0]{\setlength{\tabcolsep}{4.5pt}}
\newcommand{\tableSkip}[0]{\vspace{0.1in}}
\newcommand{\denseParActually}[1]{\textbf{#1}\hspace{1.5ex}}
\newcommand{\densePar}[1]{\paragraph{#1}}
\newcommand{\resSmall}[1]{\scalebox{0.8}{#1}}

\usepackage[american]{babel}

\usepackage{algorithm}
\usepackage{algpseudocode}
\algnewcommand\algorithmicinput{\textbf{Input:}}
\algnewcommand\algorithmicoutput{\textbf{Output:}}
\algnewcommand\Input[1]{\State\algorithmicinput\ #1}
\algnewcommand\Output[1]{\State\algorithmicoutput\ #1}
\newcommand{\algorithmicbreak}{\Statex\vspace{-5pt}\hrulefill}

\usepackage{natbib}
\usepackage{mathtools}
\usepackage{bm}
\usepackage{booktabs}
\usepackage{placeins}
\usepackage{csquotes}
\usepackage{enumitem}
\usepackage{subcaption}
\usepackage{wrapfig}
\usepackage{multirow}
\usepackage{cleveref}
\Crefname{equation}{Eq.}{Eqs.}

\newcommand{\name}[0]{GC-xLSTM}

\title{Exploring Neural Granger Causality with xLSTMs:\\Unveiling Temporal Dependencies in Complex Data}

\author{%
    \normalfont
    \textbf{Harsh Poonia}\textsuperscript{1,}\thanks{Authors contributed equally.} \qquad \textbf{Felix Divo}\textsuperscript{2,}\footnotemark[1] \qquad \textbf{Kristian Kersting}\textsuperscript{2,3,4,5} \qquad \textbf{Devendra Singh Dhami}\textsuperscript{6} \vspace{1.5ex}\\
    \textsuperscript{1}Carnegie Mellon University \quad \textsuperscript{2}AI \& ML Group, TU Darmstadt \quad \textsuperscript{3}hessian.AI \\
    \textsuperscript{4}DFKI \quad \textsuperscript{5}Centre for Cognitive Science, TU Darmstadt  \quad \textsuperscript{6}TU Eindhoven \vspace{0.5ex}\\
    \texttt{hpoonia@cs.cmu.edu} \quad \texttt{d.s.dhami@tue.nl} \\
    \texttt{\{felix.divo,kersting\}@cs.tu-darmstadt.de}
}

\begin{document}

\maketitle
\setcounter{footnote}{0} %

\begin{abstract}
Causality in time series can be challenging to determine, especially in the presence of non-linear dependencies. Granger causality helps analyze potential relationships between variables, thereby offering a method to determine whether one time series can predict---Granger cause---future values of another. Although successful, Granger causal methods still struggle with capturing long-range relations between variables. To this end, we leverage the recently successful Extended Long Short-Term Memory~(xLSTM) architecture and propose Granger causal xLSTMs~(\name{}). It first enforces sparsity between the time series components by using a novel dynamic loss penalty on the initial projection. Specifically, we adaptively improve the model and identify sparsity candidates. Our joint optimization procedure then ensures that the Granger causal relations are recovered robustly. Our experimental evaluation on six diverse datasets demonstrates the overall efficacy of \name{}.
\end{abstract}

\input{content/body}
\FloatBarrier

\begin{ack}%
    \vspace{-1ex}
    This work received funding from the ACATIS Investment KVG mbH project \enquote{Temporal Machine Learning for Long-Term Value Investing} and the KompAKI project of the German Federal Ministry of Research, Technology and Space within the \enquote{The Future of Value Creation -- Research on Production, Services and Work} program~(funding number 02L19C150), managed by the Project Management Agency Karlsruhe~(PTKA).
    The TU Eindhoven author received support from their Dep. of Mathematics and Computer Science and the Eindhoven AI Systems Institute.
    Furthermore, this work benefited from the DYNAMIC Centre funded by the LOEWE program of the Hessian Ministry of Science and Research, Arts and Culture~(HMWK) as LOEWE 1/16/519/03/09.001 (0009)98 and the HMWK project \enquote{The Third Wave of Artificial Intelligence -- 3AI}.
    It also benefited from the early stage of the German Federal Ministry for Economic Affairs and Energy project \enquote{Souveräne KI für Europa}~(13IPC040G) as part of the EU funding program IPCEI-CIS; funding has not started yet,
    and from early stages of the Cluster of Excellence \enquote{Reasonable AI} funded by the German Research Foundation~(DFG) under Germany's Excellence Strategy -- EXC-3057; funding will begin in 2026.
    Funding for H.~Poonia to attend NeurIPS was provided by the CMU GSA/Provost Conference Funding.
    The authors are responsible for the content of this publication.
    Map data \copyright~OpenStreetMap contributors, licensed under the ODbL and available from \href{https://www.openstreetmap.org}{openstreetmap.org}.
\end{ack}

\clearpage
\bibliography{references}

\appendix

\clearpage
\input{content/checklist}

\clearpage
\input{content/appendix}

\end{document}

%% file: math_commands.tex
\usepackage{amsmath,amsfonts,bm}

\def\eqref#1{equation~\ref{#1}}

\def\1{\bm{1}}

\DeclareMathAlphabet{\mathsfit}{\encodingdefault}{\sfdefault}{m}{sl}
\SetMathAlphabet{\mathsfit}{bold}{\encodingdefault}{\sfdefault}{bx}{n}

\newcommand{\N}{\mathbb{N}}

\newcommand{\R}{\mathbb{R}}

\newcommand{\softmax}{\mathrm{softmax}}

\newcommand{\normltwo}{L^2}

%% file: content/body.tex
\section{Introduction}
\label{sec:intro}

Finding cause and effect among and within a group of multivariate time series can lead to a better understanding of the dynamics of the involved time series.
For instance, in computational neuroscience and medicine, discovering brain connectivity assists in better understanding natural cognition~\citep{smithNetworkModellingMethods2011}.
Discovering inter-dependencies between time series also has a critical impact on many other research areas, such as finance~\citep{masini2023machine}, climate science~\citep{mudelsee2019trend}, and industrial applications~\citep{strem2025apt}.
Although efforts have recently been made to improve the interpretability of time series models~\citep{ismail2020benchmarking,turbe2023evaluation}, most methods are restricted to finding post-hoc interpretations and only focus on short-term dependencies.

The framework of Granger causality (GC)~\citep{granger1969investigating} was introduced to address the challenge of determining whether one variable's past values can help forecast another's future values, without implying direct causality.
GC can be established using statistical hypothesis tests, determining whether one time series can predict another.
The test traditionally involves estimating a vector autoregressive model and examining whether lagged values of a time series improve or degrade the prediction of the other, while controlling the past behavior of both series.
Although GC does not imply a direct cause-and-effect relationship between the involved time series~\citep{heckman2008econometric}, recognizing these interdependencies can lead to a better understanding of the dynamic relationships between variables over time~\citep{marcinkevicsInterpretableModelsGranger2021,shojaie2022granger}.

Many families of deep learning architectures have been explored for time series analysis over the years, such as multilayer perceptrons \citep{zengAreTransformersEffective2023,dasLongtermForecastingTiDE2023}, recurrent neural networks~(RNNs) \citep{hochreiterLongShortTermMemory1997,choLearningPhraseRepresentations2014}, convolutional neural networks \citep{wuTimesNetTemporal2DVariation2022,wangMICNMultiscaleLocal2022}, Transformers \citep{vaswaniAttentionAllYou2017,nieTimeSeriesWorth2023}, state-space models~(SSMs) \citep{wangMambaEffectiveTime2025}, or mixing architectures~\citep{wangTimeMixerDecomposableMultiscale2024}.
Throughout this, recurrent models remained a natural choice for time series data since their direction of computation aligns well with the forward flow of time.
This aligns particularly well with the goals of neural Granger causality.
Although SSMs are comparable in that regard, RNNs tend to offer more powerful forecasting capabilities since they deteriorate less when modeling long-term dependencies.
Furthermore, their inference runtime is typically linear in the sequence length at constant memory cost, making them much more efficient than, for instance, Transformers with quadratic runtimes and memory requirements, while remaining highly expressive.
Recently, \citet{beckXLSTMExtendedLong2024} revisited recurrent models by borrowing insights gained from Transformers in many domains, specifically natural language processing.
Their proposed Extended Long Short-Term Memory~(xLSTM) model sparked a resurgence of interest in recurrent architectures for sequence modeling and has already proven highly suitable for time series forecasting \citep{krausXLSTMMixerMultivariateTime2025,alharthiXLSTMTimeLongTermTime2024}.

Although most Granger causal machine learning methods assume linearity in time series as a fundamental prerequisite~\citep{siggiridou2015granger,zhang2020cause}, recent efforts capture non-linear dynamics in time series by using neural networks as the modeling choice instead of VARs~\citep{tank2022neural,lowe2022amortized,cheng2024cuts}.
Although successful, these non-linear methods require careful feature engineering to include time-based patterns. Thus, they may not capture interactions between time series and external factors as effectively as xLSTMs, which can learn non-linear patterns and adapt to the non-stationary nature of time series data.

We introduce \emph{\name}, a novel method that leverages xLSTMs to uncover the GC relations in the presence of complex data, which inherently can have long-range dependencies.\footnote{Code available at \codeLink{}.}
\name{} first enforces sparsity between the time series components by using a novel lasso penalty on the initial projection layer of the xLSTM.
We learn a weight per time series and then adapt them to find the relevant variates for that step.
Then, each time series component is modeled using a separate xLSTM model, enabling us to discover interpretable GC relationships between the time series variables.
After the forecast results by the individual xLSTM models, the important features are made more prominent, whereas the less important ones are diminished by a joint optimization technique, which includes using a novel reduction coefficient.
Thus, the overall \name{} model can be trained end-to-end to uncover long-range Granger causal relations.

Our main research \textbf{contributions} can be summarized as follows:
\begin{enumerate}[label=(\roman*)]
    \item We propose \name{}, a novel model that can uncover Granger causal relations in non-linear time series.
    \item Our novel algorithm jointly improves the forecasting model while adaptively enforcing strict sparsity.
    \item Our empirical evaluations demonstrate that \name{} can robustly discover Granger causal relations in the presence of complex simulated and real-world data.
\end{enumerate}

\densePar{Outline.}
We start by recalling preliminaries and reviewing related research to contextualize this work in the broader body of research on neural Granger causality in \Cref{sec:prelim_related}.
This allows us to introduce \name{} in \Cref{sec:method} and empirically evaluate in relation to other methods in \Cref{sec:exp}.
Finally, we conclude with an outlook to future work in \Cref{sec:conclusion}.

\section{Preliminaries and Related Work}
\label{sec:prelim_related}

We are interested in datasets of strictly stationary time series $\bm S \in \R^{V \times T}$ of $V$ variates with length $T$.
Let $\bm{S}_t \in \R^V$ denote the value of $\bm S$ at time $t$.
A variate $v$ (sometimes called a channel) can be any scalar measurement, such as the chlorophyll content of a plant or the spatial location of some object being tracked.
Its value at time $t$ is $S_{v,t} \in \R$.
The measurements are assumed to be carried out jointly at $T$ regularly spaced time steps.
In forecasting, a model is presented with a time series of $C$ context steps $\bm{S}_{<t}$ before $t$, from which it shall predict the next value $\bm{S}_{t} \in \R^V$.

\subsection{Granger Causality}
\label{sec:prelim_related:gc}

If the observed time series were generated by some underlying process $g$, which we can formalize as a structural equation model for all time steps $t$ as
  $  S_{v,t} = g_v(\bm{S}_{1,<t}, \dots, \bm{S}_{V,<t}) + \epsilon_{v,t} \text{ for all } v \in \mathcal{V}$,
where $\epsilon_{v,t}$ is some additive zero mean noise independent from all variates $\mathbf{S}_{v,<t}$ and $\mathcal V := \{1,\dots,V\}$ is the set of all variates.
In Granger causality~\citep{granger1969investigating}, we aim to determine whether past values $\bm{S}_{v,<t}$ of a variate $v$ are predictive for future values $\bm{S}_{w,\geq t}$ of another variate $w$. Following the notation of \citet{shojaie2022granger}, we formally define:
\begin{definition}[Granger Causality] \label{def:GC}
    Variate $v$ is \emph{Granger non-causal} for $w$ if and only if $g_v$ is invariant to $\bm{S}_{w,<t}$ for all $t \in \{1,\dots,T\}$, i.e., if and only if
    $$
        g_v(\bm{S}_{1,<t}, \dots, \bm{S}_{V,<t}) = g_v(\bm{S}_{1,<t}, \dots, \bm{S}_{V,<t} \setminus \bm{S}_{w,<t}) .
    $$
    Else, we call $v$ \emph{Granger causal} for $w$.
\end{definition}

The set of all such relationships are the directed edges $\mathcal{E} \subseteq \mathcal{V} \times \mathcal{V}$ of the \emph{Granger causal graph} $(\mathcal{V}, \mathcal{E})$ of the variates, which we eventually aim to uncover.

\subsection{Neural Granger Causality}
\label{sec:prelim_related:ngc}

Unfortunately, however, we cannot explicitly access $g$ in most realistic settings.
Using machine learning methods, we can nonetheless estimate each of the $V$ process components $g_v$ by an autoregressive time series forecasting model $\mathcal{M}_{\bm\theta,v}(\bm{S}_{<t} ) \approx g_v(\bm{S}_{<t} ) = \bm{S}_{v,t} - \epsilon_{v,t}$.
We can do so by estimating the parameters $\bm\theta$ of the model based on the dataset of time series, minimizing the predictive mean squared error~(MSE) loss
\begin{equation}
    \mathcal{L}_\text{pred}\left(\bm\theta \right) = \sum_{v=1}^V \sum_{t=1}^T \left( S_{v,t} - \mathcal{M}_{\bm\theta,v}(\bm{S}_{<t} ) \right)^2 .
    \label{eq:loss_pred}
\end{equation}
In the case of using neural networks for $\mathcal{M}_{\bm\theta,v}$, this approach is called \emph{Neural Granger Causality}~\citep{tank2022neural}.

It would be very costly to train a total of $V^2$ models to test if each variate Granger causes any other variate and thus construct the entire Granger causal graph.
To avoid this, we can train merely a single \emph{component-wise} model $\mathcal{M}_{\bm\theta,v}$ for each variate $v$ and inspect what inputs $w$ it is sensitive to.
While this does not ablate models as in classical GC, it reduces the number of models to be trained from quadratic to linear in $V$.
This can, for instance, be achieved by optimizing the predictive loss $\mathcal{L}_\text{pred}\left(\bm\theta \right)$ based on all model parameters $\bm\theta$ with a regularizer $\Omega(\widehat{\bm{\theta}}_v )$ enforcing sparsity in the input features:
\begin{equation}
    \min_{\bm\theta, \widehat{\bm{\theta}}} \;
    \mathcal{L}_\text{pred}\left(\bm\theta \right) +
    \lambda \sum_{v=1}^V \Omega\left(\widehat{\bm{\theta}}_v \right),
    \label{eq:minimization_general}
\end{equation}
where $\widehat{\bm{\theta}}_v$ are tunable parameters of the regularizer for variate $v$ and $\lambda \in \R_+$ is a hyperparameter to adjust the degree of sparsity.
One such approach are cMLPs~\citep{tank2022neural}; multilayer perceptrons where the first weight matrix $\bm W \in \R^{D \times V}$ projecting from $V$ features at each time step to $D$ hidden dimensions is regularized to encode a sparse selection of input features.
$\Omega$ is instantiated as an $\normltwo$ norm of its columns: $\sum_{v=1}^V \norm{ \bm{W}_v }_2$. Note that $\bm\theta$ and $\widehat{\bm{\theta}}$ can overlap.

Sparsity can then be extracted by binarizing the entries of $\bm W$ using a user-defined threshold $\tau$.
This is necessary as the $\normltwo$ penalty tends only to shrink parameters to small values near zero, yet not clamp them sharply to it.
This, however, allows subsequent layers to amplify the dampened signal again and still use it for forecasting.
We avoid this disadvantage in \name{} by explicitly optimizing the feature extractor for strict sparsity.
This more principled approach works without determining a sparsity threshold $\tau$.

Previous work has explored both more regularizers \citep{tank2022neural} and different means to extract Granger causal relationships, including using feature attribution via explainability \citep{atashgahiUnveilingPowerSparse2024} and interpretability \citep{marcinkevicsInterpretableModelsGranger2021}.
Furthermore, several works have gone towards learning relevant representations that respect the underlying Granger causality~\citep{xu2016learning,varando2021learning,dmochowski2023granger}.
\citet{zoroddu2024learning} present another approach where prior knowledge is encoded in the form of a noisy undirected graph, which aids the learning of Granger causality graphs.
A more recent approach~\citep{lin2024granger} employs Kolmogorov-Arnold networks~\citep{liu2024kan} to learn the Granger causal relations between time series.
For a discussion of when Granger causality implies (Pearlian) causality, we refer to \citet{pettenuzzoGrangerCausalityExogeneity2011} and \citet{dasNonAsymptoticGuaranteesReliable2023}, allowing to connect to works such as the one of \citet{rubensteinCausalConsistencyStructural2017}.

\subsection{Extended Long Short-Term Memory (xLSTM)}
\label{sec:prelim_related:xlstm}

\citet{beckXLSTMExtendedLong2024} propose two building blocks to build up xLSTM architectures: the sLSTM and mLSTM modules for vector-valued (i.e., multivariate) sequences.
sLSTM cells improve upon classic LSTMs by exponential gating.
For parallelizable training, mLSTM cells replace memory mixing between hidden states with an associative matrix memory.
We will continue by recalling how sLSTM cells function since we found their memory mixing more effective in time series forecasting.

The standard LSTM architecture of \citet{hochreiterLongShortTermMemory1997} updates the cell state $\mathbf{c}_t$ through a combination of input, forget, and output gates, which regulate the flow of information across tokens.
sLSTM blocks, owing to the contained sLSTM cells, enhance this by incorporating exponential gating and memory mixing~\citep{greffLSTMSearchSpace2017} to better handle complex temporal and cross-variate dependencies.
Additional normalization states are introduced to stabilize training under the new exponential activation function.
As \citeauthor{beckXLSTMExtendedLong2024} have shown, it is sufficient and computationally beneficial to constrain the memory mixing performed by the recurrent weight matrices $\bm{R}_z, \bm{R}_i, \bm{R}_f,$ and $\bm{R}_o$ to individual \emph{heads}.
This is inspired by the multi-head setup of Transformers~\citep{vaswaniAttentionAllYou2017}, yet more restricted and efficient.
In particular, each token gets broken up into groups of features, where the input weights $\bm{W}_{z,i,f,o}$ act across all of them, but the recurrence matrices $\bm{R}_{z,i,f,o}$ are implemented as block-diagonal.
This permits specialization of the individual heads to patterns specific to the respective section of the tokens and empirically does not sacrifice expressivity.

\begin{figure*}[t]
    \centering
    \includegraphics[width=\linewidth]{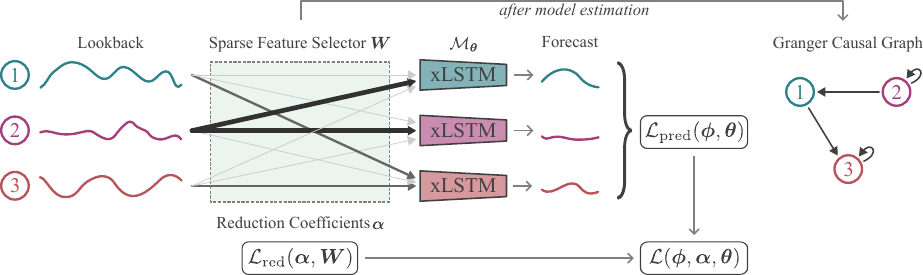}
    \caption{\textbf{\name{} performs three key steps to determine the Granger causal links:} Firstly, for each time series component, all variates are embedded with a sparse feature encoder $\bm W$ that is regularized through a novel sparsity loss with learned reduction coefficients $\bm\alpha$. xLSTM models then learn to autoregressively predict future steps from that embedding. Finally, once model estimation is complete, Granger causal dependencies can be extracted from $\bm W$.}
    \label{fig:architecture}
\end{figure*}

\section{\name{}}
\label{sec:method}

We will now introduce the \name{} architecture and detail the optimization for strict sparsity jointly with the model parameters.
At the end, we will additionally discuss theoretical properties of the proposed system.

\subsection{Overall Architecture}
\label{sec:method:overall}

As \Cref{fig:architecture} shows, we estimate a pipeline of sparse feature selectors and xLSTM models to predict the multivariate time series.
Eventually, this allows us to derive Granger causal dependencies from the selected features.
Specifically, we learn for each variate $v$ a separate sparse projection and compute $\bm{x}_v = \bm{W}_v \bm{S} + \bm{b}_v$ shared across time.
The matrix $\bm{W}_v \in \R^{D \times V}$ is shared across all lags of the time series for simplicity.
We will lift this restriction in \Cref{sec:exp:model_analysis}.
Note that $\bm b$ does not affect the sparse use of inputs.
We write $\bm\phi$ for the set of parameters $\bm{W}_v$ and $\bm{b}_v$ for all $v \in \mathcal{V}$.
In addition to selecting dependencies, the sparse projection embeds the data into $D$-dimensional hidden space.

For Neural Granger Causality to successfully and faithfully extract the proper underlying dependencies, it is essential to employ models that can capture the complete set of dependencies.
We, therefore, employ powerful deep-learning models with significantly higher capacity than the cMLPs and cLSTMs in prior work \citep{tank2022neural}.
In particular, we instantiate the individual time series forecasters $\mathcal{M}_{\bm\theta_v,v}$ with sLSTM blocks as introduced in \Cref{sec:prelim_related:xlstm}.
They can capture long-range dependencies in time series data, substantially enhancing the capabilities of traditional LSTMs in handling extended contexts.
\name{} consists of $V$ sLSTM models, each modeling a different time series component.
They are trained using established forecasting losses, such as the MSE loss $\mathcal{L}_\text{pred}\left(\bm\phi, \bm\theta \right)$ from \Cref{eq:loss_pred}.

\subsection{Optimizing for Strict Input Sparsity}
\label{sec:method:sparsity}

\begin{figure}[t]
    \centering
    \begin{subfigure}[t]{0.48\textwidth}
        \centering
        \includegraphics[width=\linewidth]{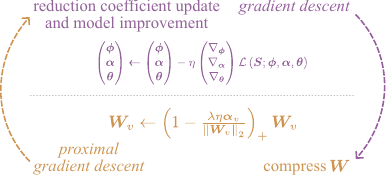}
        \caption{\textbf{The optimization procedure for \name{} alternates between \textcolor[HTML]{926299}{GD} and \textcolor[HTML]{CE9552}{proximal GD}.} A high value of $\alpha_v^w$ entails a strong compression of column $\bm{W}_v$, i.e. more sparsity in depending on variate $w$.}
        \label{fig:main_method}
    \end{subfigure}
    \hfill
    \begin{subfigure}[t]{0.48\textwidth}
        \centering
        \includegraphics[width=0.75\linewidth]{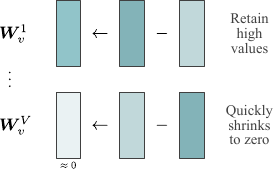}
        \caption{\textbf{Visual representation of the compression step in \Cref{alg:main}, \cref{alg:main:prox}.}
        The matrix preserves columns with high norm that are important for Granger causality detection  (dark) while suppressing others (light).}
        \label{fig:compression_step_intuition}
    \end{subfigure}
    \caption{\textbf{Optimization procedure and compression intuition for \name{}.}}
    \label{fig:combined}
\end{figure}

The purpose of the feature selector $\bm W$ is to only \enquote{pay attention} to as many variates as necessary for successful forecasting.
A common approach to achieve this sparsity on the Granger causal relationships is via the lasso regularization explained in \Cref{sec:prelim_related:ngc}.
We use a variation of the group lasso penalty \citep{yuan2006model,simonStandardizationGroupLasso2012} on the initial projection layer of \name{} as a structured sparsity-inducing penalty that encourages the selection of entire groups of variables, encoded as the columns of $\bm{W}_v$.
Note that our penalty differs from adaptive lasso~\citep{yuan2006model}, where the weights are not learned but are treated as fixed heuristics.
Note that standard gradient descent methods cannot optimize such a penalty directly due to its non-differentiability.
We thus adaptively compress by learning a reduction coefficient $\bm{\alpha}_v \in \R^V$ that selects which of the $V$ columns of $\bm{W}_v$ are redundant.
We perform this compression of $\bm{W}_v$ in a joint procedure with the general optimization of the forecasting model, as provided in \Cref{alg:main}.
Specifically, we perform two updates per optimization step for each of the variates, as \Cref{fig:main_method} depicts.
Firstly, we optimize the projection weights $\bm\phi$, the reduction coefficients $\bm\alpha$, and the xLSTM parameters $\bm\theta$ using mini-batch gradient descent.
This corresponds to \crefrange{alg:main:gradient_start}{alg:main:gradient_end} in \Cref{alg:main}.
It optimizes the following loss expected over the time series data $\bm S$:
\begin{equation} \label{eq:loss_gd}
    \min_{\bm{\phi}_v, \bm{\alpha}_v, \bm{\theta}_v} \mathcal{L}_\text{pred}(\bm S ; \bm{\phi}_v, \bm{\theta}_v) +
    \underbrace{
        \lambda \log\left( \sum_{w=1}^V \alpha_v^w \norm{\sg\left( \bm{W}_v^w \right)}_2 \right)
    }_{\mathcal{L}_\text{red}\left(\bm{\alpha}_v, \bm{W}_v \right)}
\end{equation}
Note that we, crucially, only descend on the reduction coefficient $\bm\alpha$ in $\mathcal{L}_\text{red}$ and not on $\bm{W}_v$, as the stop-gradient $\sg(\cdot{})$ denotes.
This sparsity optimization is instead performed by the second step in the procedure, shown in \cref{alg:main:prox}, where a proximal gradient descent step dynamically shrinks $\bm{W}_v$ proportional to $\bm{\alpha}_v$.
The compression update takes a descent step towards the gradient of
\begin{equation} \label{eq:loss_prox_gd}
    \lambda \sum_{w=1}^V \alpha_v^w \norm{ \bm{W}_v^w }_2 
\end{equation}
followed by a soft thresholding.
Intuitively, the $\mathcal{L}_\text{red}$ component of \Cref{eq:loss_gd} keeps $\bm{W}_v$ fixed while learning $\bm{\alpha}_v$, and \Cref{eq:loss_prox_gd} keeps $\bm{\alpha}_v$ fixed in the proximal step to compress $\bm{W}_v$.
\Cref{fig:compression_step_intuition} depicts the intuition of the proximal gradient step and soft-thresholding in \cref{alg:main:prox}.

\densePar{Details on learning the reduction loss $\mathcal{L}_\text{red}$.}
It is worth briefly discussing the use of the logarithm in \Cref{eq:loss_gd}.
It mainly gives more equal weight to the decreases in $\bm{W}_v$ column norms and encourages learning of better sparse Granger causal relations.
It furthermore normalizes the gradient updates to $\bm{\alpha}_v$.
Empirically, this loss engineering allowed training models that were significantly more robust to noise and changes to the sparsity hyperparameter $\lambda$.
This was reflected by a more stable variable usage and predictive loss $\mathcal{L}_\text{pred}$.

\densePar{Ensuring non-negativity of the reduction coefficients $\bm\alpha$.}
For the proximal update step to be well-behaved, we need to ensure that $\bm{\alpha}_v$ results in a convex combination of column weights, i.e., that $\bm{\alpha}_v^w > 0$ for all $w \in \mathcal{V}$ and $\bm{\alpha}_v^T \mathds{1} = 1$.
We achieve this by re-parameterizing it as $\bm{\alpha}_v = \softmax\left( \bm{\beta}_v \right)$, and learning $\bm{\beta}_v$ instead of $\bm{\alpha}_v$.

\densePar{Intuitive dynamics of the gradient update step.}
The weights $\bm{W}_v$ in the reduction loss $\mathcal{L}_\text{red}$ of \Cref{eq:loss_gd} only serve to learn good reduction coefficients $\bm{\alpha}_v$, and are not optimized themselves.
Deriving the gradient of the penalty term $\mathcal{L}_\text{reg}$ with respect to the underlying $\bm\beta$ provides a helpful intuition of the training dynamics:
\begin{equation*}
    \frac{\partial}{\partial \beta_v^w } \sum_{w=1}^V \alpha_v^w \norm{ \bm{W}_v^w }_2
    = \frac{\partial}{\partial {\beta_v^w} } \sum_{w=1}^V \softmax(\bm{\beta}_v)^w \norm{\bm{W}_v^w}_2
    = \alpha_v^w \hspace{-3pt} \left( \hspace{-2pt} \norm{\bm{W}_v^w}_2 - \sum_{w=1}^V \alpha_v^w \norm{ \bm{W}_v^w }_2 \hspace{-2pt} \right)
\end{equation*}
\begin{equation*}
    \Longrightarrow
    \frac{\partial}{\partial \beta_v^w } \mathcal{L}_\text{reg}
    = \lambda \frac{\partial}{\partial \beta_v^w } \log\left( \sum_{w=1}^V \alpha_v^w \norm{ \bm{W}_v^w }_2 \right)
    = \lambda \alpha_v^w \left( \frac{\norm{\bm{W}_v^w}_2}{\sum_{w=1}^V \alpha_v^w \norm{ \bm{W}_v^w }_2}  - 1\right)
\end{equation*}
We can see that if the norm of a column $\norm{\bm{W}_v^w}_2$ is large, that corresponding $\frac{\partial}{\partial \beta_v^w } \mathcal{L}_\text{reg}$ will be large.
Gradient descent will thus decrease $\alpha_v^w$ and effectively allocate less weight to its removal in the compression step.
Furthermore, $\frac{\partial}{\partial \beta_v^w } \mathcal{L}_\text{reg}$ also scales with $\alpha_v^w$, resulting in a self-reinforcing loop that aids learning sparse representations.

\densePar{Practical considerations.}
Furthermore, we perform staged optimization of $\bm\alpha$, which is initialized to a uniform distribution by setting all $\bm\beta = 0$.
We only start training $\bm\alpha$ after exploring the prediction loss and having obtained a reasonably compressed forecaster, which is controlled by the hyperparameter $K$ in \Cref{alg:main} (see \cref{alg:main:staged_alpha}).
While we present the method with mini-batch gradient descent for conciseness, modern optimizers, such as Adam~\citep{kingmaAdamMethodStochastic2017}, can further improve convergence.

\subsection{Theoretical Analysis}
\label{sec:method:theoretical_analysis}

Ultimately, we want to ensure that, provided real-world data, we can find hyperparameters such that \Cref{alg:main} discovers all and only those edges of the unique underlying GC graph $(\mathcal{V}, \mathcal{E})$ as per \Cref{def:GC}.
Providing convergence guarantees in full generality is notoriously hard for such practical architectures and optimization schemes, and thus rarely attempted.
However, we can at least investigate whether the chosen model class containing $\mathcal{M}_{\bm{\theta},v}$ can approximate $g_v$ to arbitrary precision.
If that is the case, we can be reasonably sure that gradient-based optimization schemes will yield satisfactory approximations, even without formal guarantees.

The forecasting component of \name{} consists of two main steps: the sparse initial projection $\bm{W}_v$ and the subsequent xLSTM blocks.
One might think that the sparsity of $\bm{W}_v$ hinders learning the correct $g_v$. 
However, the true underlying $g_v$ is independent of all variates $w$ without ingoing edges into variate $v$.
Thus, depending on an appropriate choice of the sparsity hyperparameter $\lambda$, the projection $\bm{W}_v$ can encode exactly those as zero entries.
It remains to investigate the approximation capabilities of the sLSTM blocks, which we present in \Cref{sec:app:xlstm_approx} in more detail.
In summary, we can assume that sLSTM blocks are at least as powerful as RNNs, which are, in turn, universal function approximators.
Thus, the overall \name{} architecture is sufficiently rich to model $g_v$ to adequate precision.
We continue by confirming this empirically in the next section.

\newpage

\section{Experimental Evaluation}
\label{sec:exp}

\begin{table*}[t]
    \centering
    \caption{\textbf{\name{} is highly accurate at discovering GC relations in the chaotic and non-linear Lorentz-96 system.} For each setting and baseline, we provide the accuracy~(Acc.), balanced accuracy~(BA), and AUROC. The best models are highlighted as \textbf{bold}.}
    \label{tab:auroc_lorenz}
    \denseColumns
    \begin{tabular}{lccccccc}
        \toprule
        & & \multicolumn{3}{c}{$F=10$} & \multicolumn{3}{c}{$F=40$}
        \\
        \cmidrule(lr){3-5} \cmidrule(lr){6-8}
        Model & Year & Acc. (\textuparrow) & BA (\textuparrow) & AUROC (\textuparrow)  & Acc. (\textuparrow) & BA (\textuparrow) & AUROC (\textuparrow) \\
        \midrule
        VAR          & \citeyear{marcinkevicsInterpretableModelsGranger2021} & 91.8\resSmall{±1.2} & 83.8\resSmall{±1.6} & 94.0\resSmall{±1.6} & 86.4\resSmall{±0.8} & 58.5\resSmall{±1.7} & 74.5\resSmall{±4.7} \\
        cLSTM        & \citeyear{tank2022neural} & 97.0\resSmall{±1.0} & 95.0\resSmall{±2.8} & 95.8\resSmall{±2.6} & 84.4\resSmall{±1.2} & 65.6\resSmall{±3.7} & 66.1\resSmall{±3.8} \\
        cMLP         & \citeyear{tank2022neural} & 97.2\resSmall{±0.5} & 95.6\resSmall{±1.6} & 96.3\resSmall{±1.8} & 68.3\resSmall{±2.7} & 80.5\resSmall{±1.7} & \textbf{97.9}\resSmall{±1.6} \\
        GC-KAN       & \citeyear{lin2024granger} & --                  & --                  & 92.1\resSmall{±0.3} & --                  & --                  & 87.1\resSmall{±0.4} \\
        TCDF         & \citeyear{nautaCausalDiscoveryAttentionBased2019} & 87.1\resSmall{±1.2} & 70.9\resSmall{±4.4} & 85.7\resSmall{±2.7} & 77.5\resSmall{±2.3} & 62.2\resSmall{±3.0} & 67.9\resSmall{±3.1} \\
        eSRU         & \citeyear{khannaEconomyStatisticalRecurrent2020} & 96.6\resSmall{±1.1} & 95.1\resSmall{±2.0} & 96.3\resSmall{±2.0} & 86.7\resSmall{±0.9} & 88.6\resSmall{±1.4} & 93.4\resSmall{±2.1} \\
        GVAR         & \citeyear{marcinkevicsInterpretableModelsGranger2021} & 98.2\resSmall{±0.3} & 98.2\resSmall{±0.6} & \textbf{99.7}\resSmall{±0.1} & 94.5\resSmall{±1.0} & 88.5\resSmall{±4.6} & 97.0\resSmall{±0.9} \\
        \textbf{\name{}} & \textbf{ours} & \textbf{99.1}\resSmall{±0.2} & \textbf{98.5}\resSmall{±1.0} & 99.3\resSmall{±0.3} & \textbf{96.3}\resSmall{±0.3} & \textbf{96.6}\resSmall{±0.3} & 88.0\resSmall{±0.2} \\
        \bottomrule
    \end{tabular}
\end{table*}

We conduct extensive experiments on six datasets to assess the practical effectiveness of \name{}.
We will now explain the chosen architecture and parameters used to train \name{} and then discuss the datasets used before presenting our obtained results in detail.
The results are split into investigating the general GC modeling capabilities on a diverse range of applications~(\Cref{sec:exp:main}) and a subsequent model analysis, including an ablation study and using different numbers of variates~(\Cref{sec:exp:model_analysis}).

\densePar{Architecture Details.}
For our component-wise networks, we use a single xLSTM block comprised of one sLSTM layer, followed by a linear layer to predict the next time step from the preceding $C=10$ time steps~(see \Cref{tab:dataset_details} in \Cref{sec:app:datasets} for deviations).
We then directly optimize the architecture on each time step.
The hidden dimension of the sLSTM block is set to 32 for all datasets.
We find that the presence of a gated MLP to up- and down-project the hidden states of the sLSTM block does not significantly improve performance, so we omit it in all our experiments for simplicity.
We deliberately do not use any mLSTM blocks, as we also find that the sLSTM blocks are superior at capturing long-range dependencies in the data~\citep{krausXLSTMMixerMultivariateTime2025}.

\densePar{Training and Evaluation Details.}
We use the Adam optimizer~\citep{kingmaAdamMethodStochastic2017} for full gradient descent training with a weight decay of 0.1.
We schedule the learning rate to follow a linear warmup of 2,000 iterations to $\eta = 10^{-4}$, followed by cosine annealing until the end of training for a total of 13,000 steps.
We start learning the reduction coefficients after a warmup period of $K=1,500$ iterations, during which the uniform compression across all columns combined with the prediction loss gives reasonable priors for the gradient directions of the reduction coefficients.
Due to the moderate size of the datasets, we performed full-batch gradient descent.
Only the sparsity hyperparameter $\lambda$ was tuned specifically for each setting.
This allows obtaining degrees of sparsity specifically tailored to the characteristics and requirements of each dataset and task.
Note that, except for the customization of $\lambda$, we use essentially the same hyperparameter configuration for an extensive set of datasets, underpinning the robustness of \name{}.
Following \citep[Sec.~6.1]{tank2022neural}, we compute all AUROC scores by sweeping over $\lambda \in \{5, \dots, 15\}$ in steps of one.
This effectively integrates this hyperparameter as it covers the entire empirically viable range.
All training runs were carried out on a single NVIDIA RTX A6000 GPU and concluded in at most 1.5 hours.
Mini-batching could further decrease this modest training time.

\densePar{Datasets.}
We evaluate GC detection with \name{} on six diverse datasets.
Obtaining objective truth about the underlying graph for real-world scenarios is a constant challenge in Granger causality research.
We thus employ the Lorenz-96 system of differential equations~\citep{Karimi_2010}, realistic fMRI brain activity simulations~\citep{smithNetworkModellingMethods2011}, and simulated linear VAR data following \citet{tank2022neural}.
For further qualitative insights on real-world data, we additionally analyze the Moléne weather dataset~\citep{molene2015dataset}, human motion capture recordings~\citep{cmu_mocap}, and company fundamentals~\citep{divoForecastingCompanyFundamentals2025}.
An overview and more details are provided in \Cref{sec:app:datasets}.

\begin{figure*}[t]
    \centering
    \newcommand{\widthMolene}[0]{0.7\linewidth}
    \begin{subfigure}[t]{0.49\linewidth}
        \centering
        \includegraphics[width=\widthMolene]{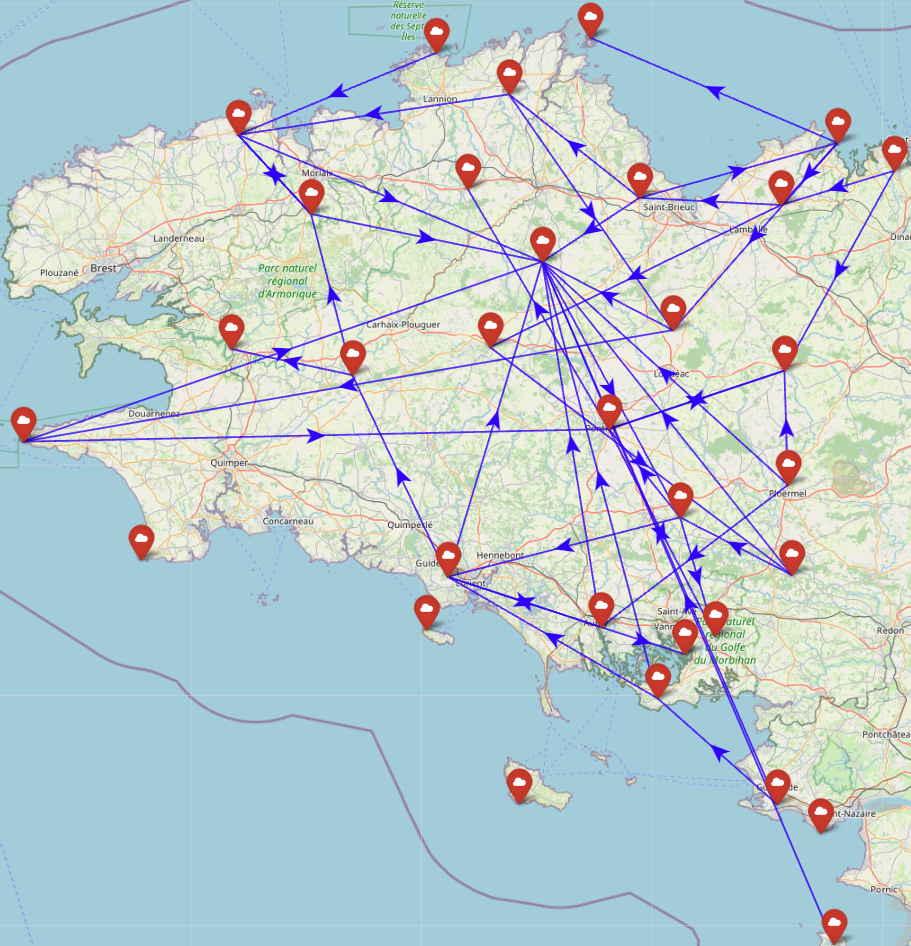}
        \caption{$\lambda = 8$.}
        \label{fig:molene_dense}
    \end{subfigure}
    \hfill
    \begin{subfigure}[t]{0.49\linewidth}
        \centering
        \includegraphics[width=\widthMolene]{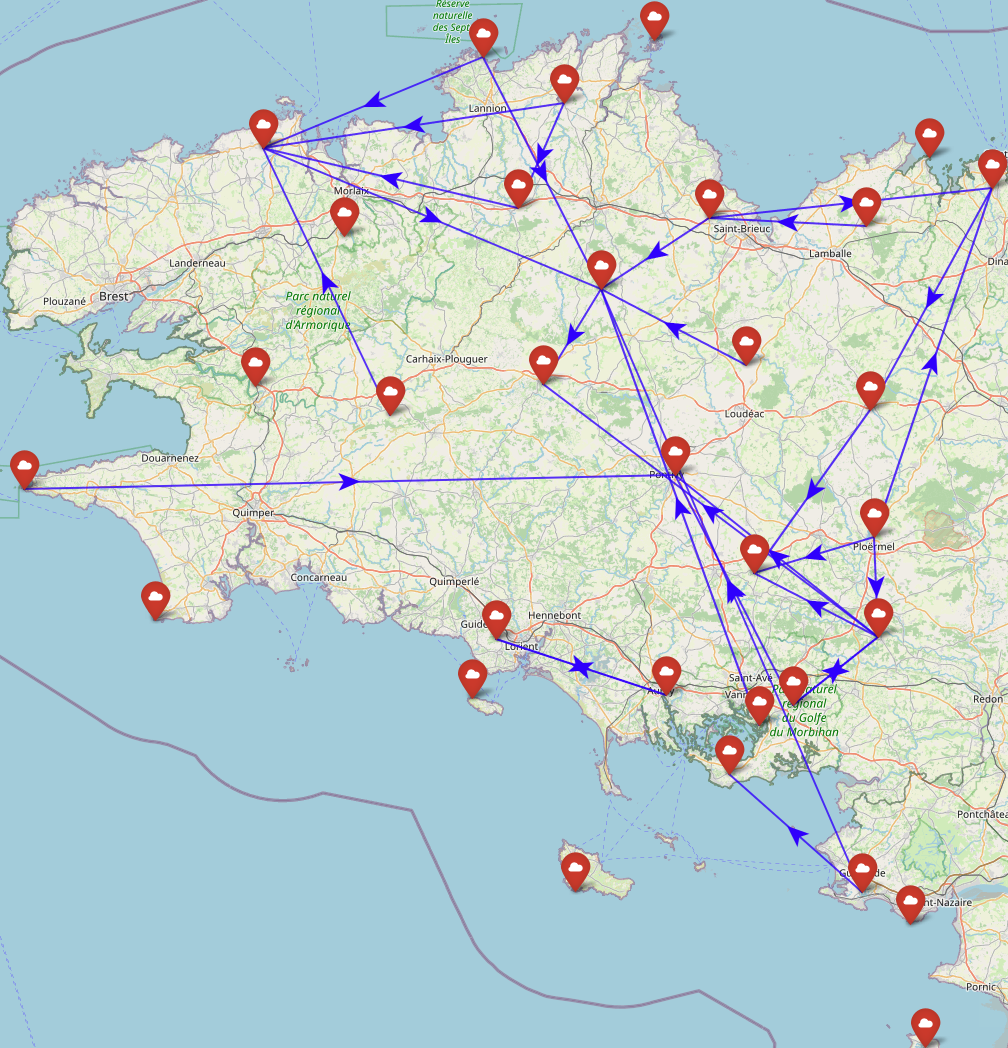}
        \caption{$\lambda = 10$.}
        \label{fig:molene_sparse}
    \end{subfigure}
    \caption{\textbf{\name{} uncovers dynamic GC weather patterns in the Moléne dataset.} We observe that the sparsity of the learned Granger causal relations increases with higher $\lambda$.}
    \label{fig:_molene_combined}
\end{figure*}

\begin{figure*}[t]
    \centering
    \newcommand{\widthMoCap}[0]{0.24\linewidth}
    \begin{subfigure}[t]{\widthMoCap}
        \centering
        \includegraphics[width=\linewidth]{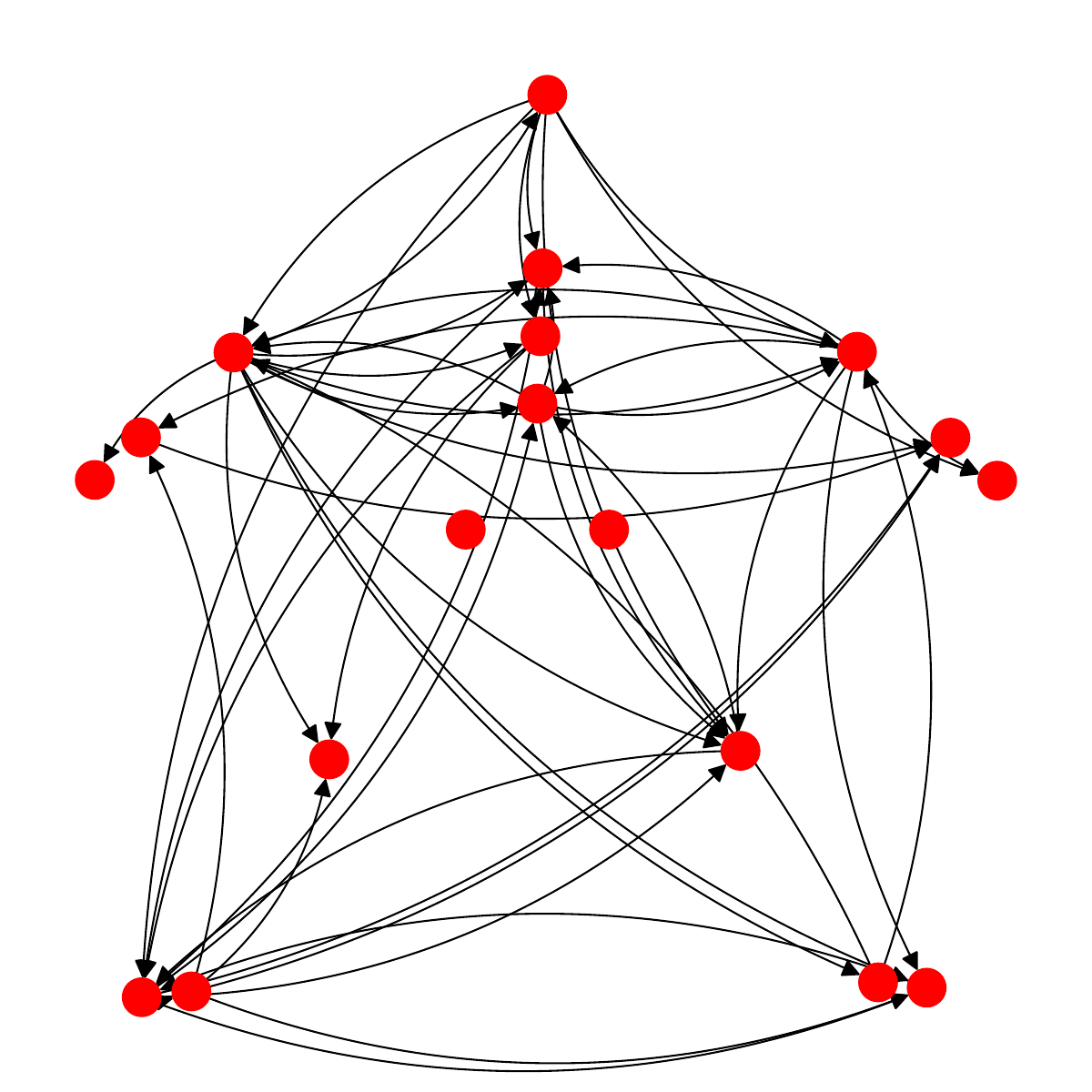}
        \caption{Salsa with $\lambda = 4$.}
        \label{fig:salsa_A}
    \end{subfigure}
    \hfill
    \begin{subfigure}[t]{\widthMoCap}
        \centering
        \includegraphics[width=\linewidth]{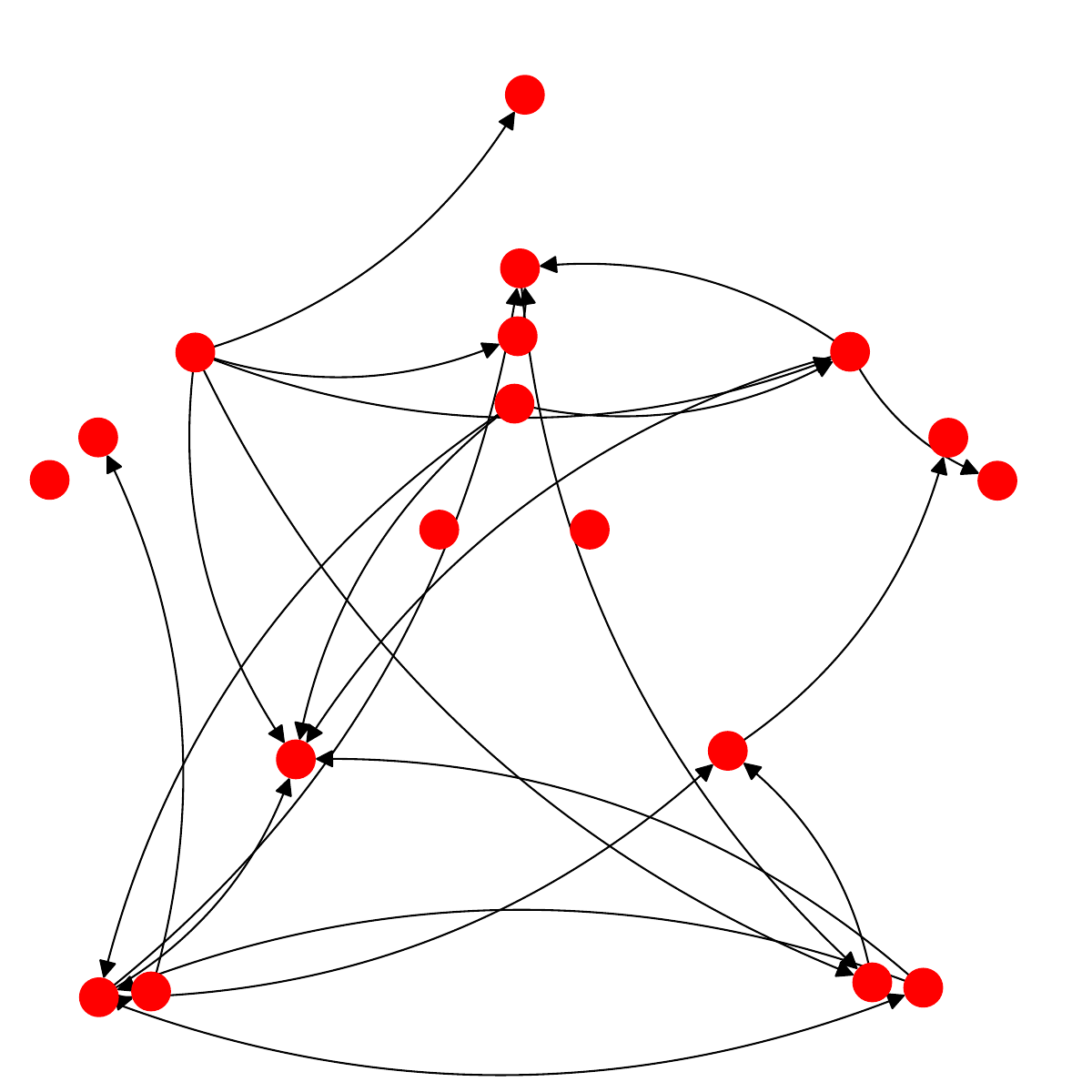}
        \caption{Salsa with $\lambda = 5$.}
        \label{fig:salsa_B}
    \end{subfigure}
    \hfill
    \begin{subfigure}[t]{\widthMoCap}
        \centering
        \includegraphics[width=\linewidth]{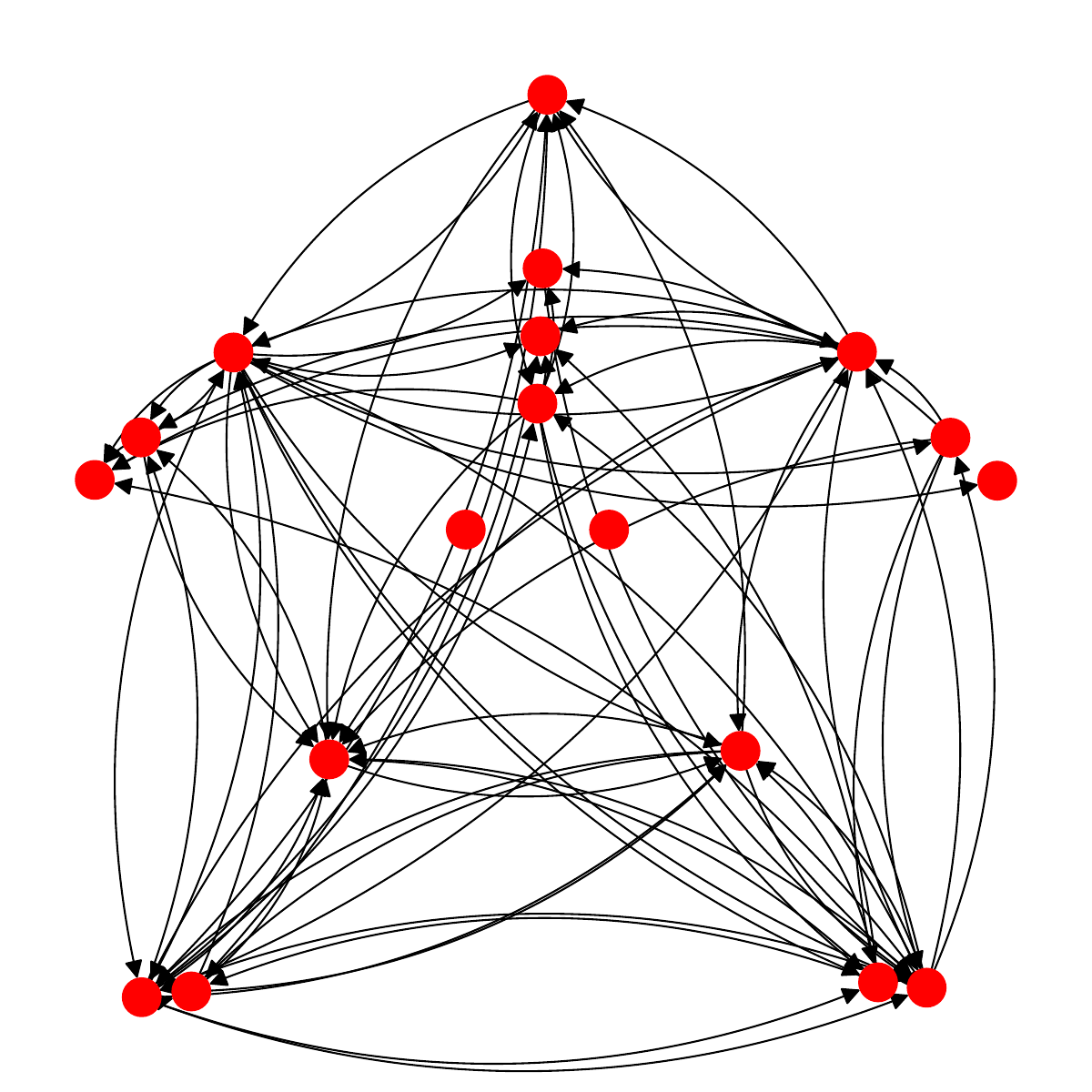}
        \caption{Running with $\lambda = 4$.}
        \label{fig:running_A}
    \end{subfigure}
    \hfill
    \begin{subfigure}[t]{\widthMoCap}
        \centering
        \includegraphics[width=\linewidth]{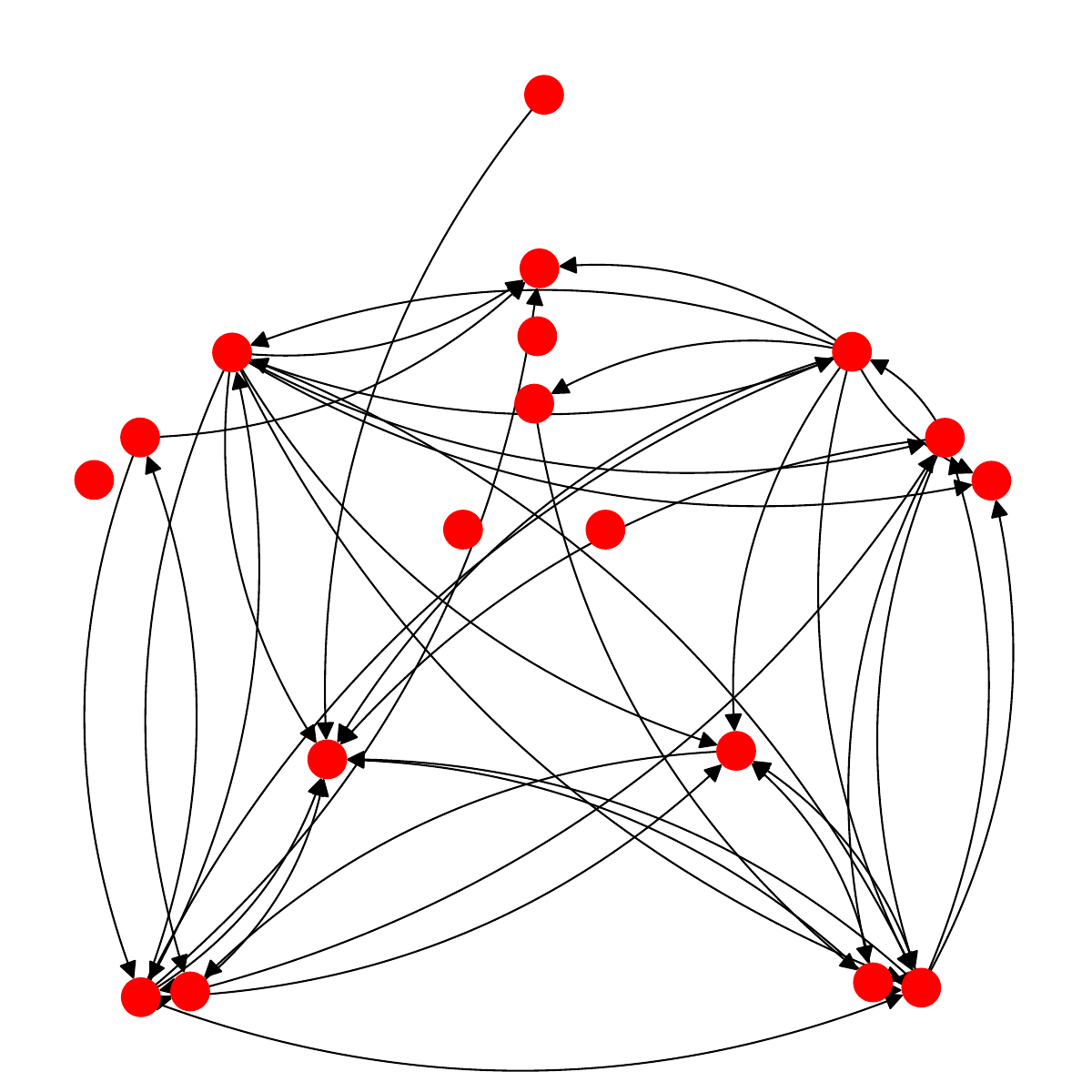}
        \caption{Running with $\lambda = 6$.}
        \label{fig:running_B}
    \end{subfigure}
    \caption{\textbf{\name{} captures complex human motions.} \name{} is able to uncover complex real-world dependencies in the Human Motion Capture dataset, giving us an intuitive understanding of the learned interactions.}
    \label{fig:mocap}
\end{figure*}

\subsection{Main Results}
\label{sec:exp:main}

\begin{wrapfigure}{r}{0.5\textwidth}
    \centering
    \vspace{-13.3pt}
    \includegraphics[width=\linewidth]{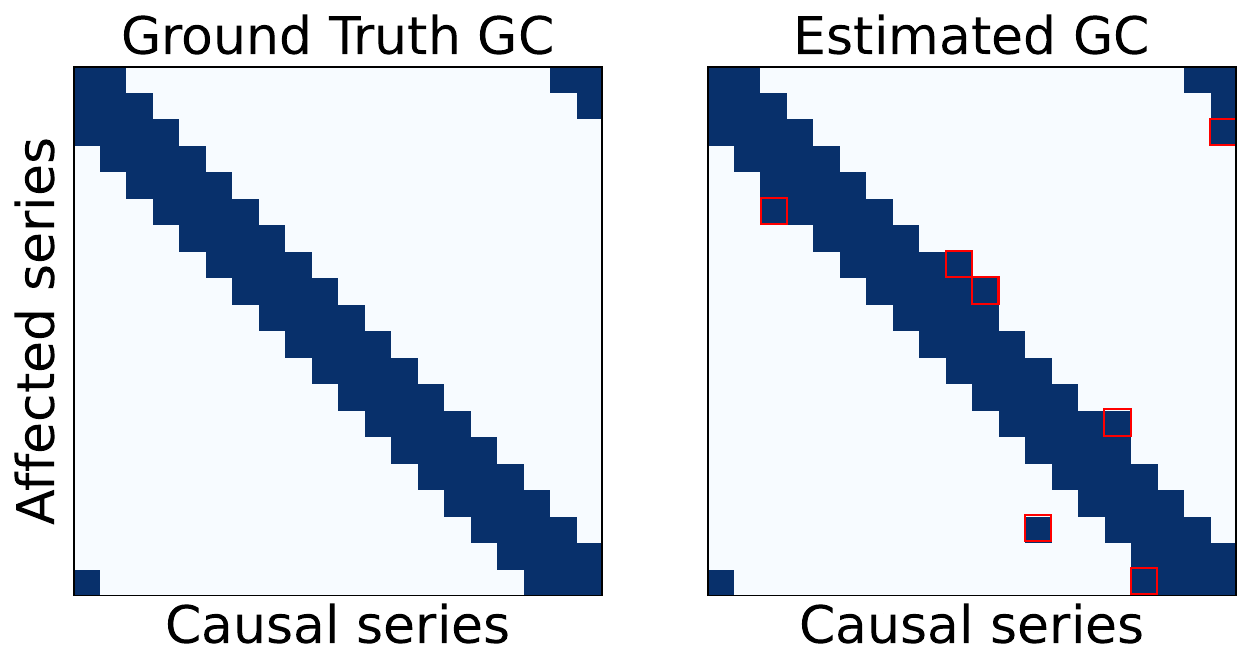}
    \caption{\textbf{\name{} uncovers the vast majority of GC edges.} In the highly chaotic $F = 40$ setting of the Lorenz-96 system \name{} is accurate in predicting the GC edges, shown in dark blue~\textcolor[RGB]{0,0,80}{\rule{0.875em}{0.875em}}. Errors are marked red~\textcolor{red}{\raisebox{0.65ex}{\fbox{\phantom{\rule{0.4ex}{0.4ex}}}}}.}
    \label{fig:lorenz96-accuracy}
    \vspace{-\baselineskip}
\end{wrapfigure}
\densePar{Lorenz-96.}
\Cref{tab:auroc_lorenz} evaluates GC detection with \name{} on three metrics.
As a qualitative comparison, we provide results for seven well-known baselines: VAR as classic $F$-tests for Granger causality taken from \citet{marcinkevicsInterpretableModelsGranger2021}, cMLP and cLSTM~\citep{tank2022neural}, GC-KAN~\citep{lin2024granger}, TCDF~\citep{nautaCausalDiscoveryAttentionBased2019}, eSRU~\citep{khannaEconomyStatisticalRecurrent2020}, and GVAR~\citep{marcinkevicsInterpretableModelsGranger2021}.
Their scores are taken as reported in the original papers. 
First, \name{} outperforms all baseline methods for both $F = 10$ and $F = 40$ in both accuracy and balanced accuracy.
This shows that \name{} reliably captures the underlying Granger causal relationships in the presence of limited and noisy data.
Second, it demonstrates that the sparsity hyperparameter $\lambda$ is rather well-behaved, as the AUROC score resulting from a sweep over a range of $\lambda$ values still provides very competitive results.
Third, as a qualitative validation, \Cref{fig:lorenz96-accuracy} visually confirms the strong prediction accuracies of \name{}.

\densePar{Moléne.}
Unlike \citet{zoroddu2024learning}, who incorporate graph prior knowledge based on sensor locations, our approach learns the GC structure solely from the temperature observations.
This ensures that \name{} does not inherently favor regional connections over long-range dependencies, allowing it to discover dominant weather patterns operating both locally and across broader spatial scales.
Adjusting $\lambda$ allows balancing granularity and interpretability for insights into both local and regional dependencies.
The dense structure of the resulting \Cref{fig:molene_dense} exhibits a richer set of GC interactions, while the more sparse \Cref{fig:molene_sparse} highlights only the most pronounced edges.

\begin{wraptable}{r}{0.26\textwidth}
    \centering
    \vspace{-13.5pt}
    \caption{\textbf{\name{} discovers brain connectivity highly accurately.}}
    \label{tab:fmri_bold}
    \begin{tabular}{l@{\hspace{6pt}}c}
        \toprule
        Model & BA (\textuparrow) \\ %
        \midrule
        TCDF    & 72.8\resSmall{±6.3} \\
        GVAR    & 65.2\resSmall{±4.5} \\
        VAR     & 51.3\resSmall{±1.5} \\
        cMLP    & 61.4\resSmall{±6.8} \\
        cLSTM   & 65.5\resSmall{±5.3} \\
        GC-xLSTM & \textbf{73.3}\resSmall{±3.0} \\
        \bottomrule
    \end{tabular}
    \vspace{-10pt}
\end{wraptable}
\densePar{fMRI.}
Next, we evaluated the efficacy of \name{} in noisy settings by considering the rich and realistic BOLD deconvolved time series provided by the fMRI data.
As the balanced accuracies~(BA) in \Cref{tab:fmri_bold} show, \name{} significantly outperforms the baseline models.

\densePar{Human Motion Capture.}
To analyze GC relations between body joints, we focus on two specific activities: Salsa dancing and running, which provide interpretable motion patterns.
\Cref{fig:mocap} shows the results of the learned graphs for those activities. 
A closer look offers an intuitive understanding of the learned interactions.
For example, in the Salsa dance, we observe edges from the feet to the knees and the arms, supporting the characteristic movements of the lower driving the upper body.
We can also see the cross-limb correlation, with movements initiating on one side of the body and propagating across.
Similarly, the results for running strongly establish the lower limbs as primary motion drivers, with edges from the feet and knees to the arms.
The cyclic dependencies between the knees, ankles, and feet capture the repetitive, alternating nature of the gait.

\begin{wrapfigure}{r}{0.45\textwidth}
    \centering
    \vspace{-13.3pt}
    \includegraphics[width=\linewidth]{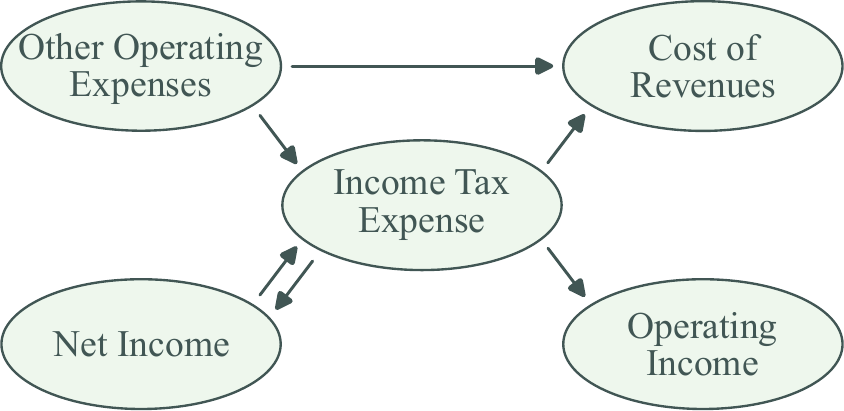}
    \caption{\textbf{\name{} extracts relations between company fundamentals.}}
    \label{tab:company_fundamentals_zoomed}
    \vspace{-2\baselineskip}
\end{wrapfigure}
\densePar{Company Fundamentals.}
Lastly, we evaluated how well \name{} can uncover relationships between financial indicators of large companies.
The dataset consists of short time series of quarterly performance indicators of 2527 large publicly traded companies.
As the excerpt in \Cref{tab:company_fundamentals_zoomed} shows and as verified by a financial expert, those extracted edges are mostly economically sensible.
Results on the full set of 19 features are provided in \Cref{sec:app:extended_exp}.

\subsection{Model Analysis}
\label{sec:exp:model_analysis}

\densePar{Ablation Study.}
\name{} comprises two key innovations: Employing the xLSTM architecture~(cf. \Cref{sec:method:overall}) and a novel joint optimization strategy~(cf. \Cref{sec:method:sparsity}).
To disentangle the empirical impact of each, we performed an ablation study.
\Cref{tab:ablation_results} to the right shows balanced accuracy results on the Lorentz ($F=40$) and fMRI datasets.
\begin{wraptable}{r}{0.65\textwidth}
    \centering
    \vspace{-13.3pt+0.5\baselineskip}
    \caption{\textbf{Both the xLSTM forecaster and the novel joint optimization of \name{} drive its improved performance.}}
    \label{tab:ablation_results}
    \denserColumns
    \begin{tabular}{ccccc}
        \toprule
        Ablation & Forecaster & Optimization & Lorentz & fMRI \\
        \midrule
        \name{} & xLSTM & Joint & \textbf{96.6\resSmall{±0.3}} & \textbf{73.3\resSmall{±3.0}} \\
        (I)     & \phantom{x}LSTM  & Joint & 93.0\resSmall{±0.3} & 62.8\resSmall{±2.0} \\
        (II)    & xLSTM & Group Lasso & 73.0\resSmall{±4.6} & 65.4\resSmall{±2.0} \\
        \bottomrule
    \end{tabular}
\end{wraptable}
First, to assess the architectural contribution, we replaced the xLSTM block with a standard LSTM.
The resulting substantial performance drop (\name{} \textrightarrow{} (I)) validates the importance of the xLSTM's advanced modeling capacity.
Second, to evaluate the optimization strategy, we substituted it with standard Group Lasso~\citep{simonStandardizationGroupLasso2012}.
This also led to a marked decline in performance  (\name{} \textrightarrow{} (II)), confirming its ability to enforce strict input sparsity.

\begin{wraptable}{r}{0.5\textwidth}
    \centering
    \vspace{-13.3pt}
    \caption{\textbf{\name{} discovers specific relations per lag.} Balanced Accuracies for simulated VAR at different lengths $T$ and number of variates $V$.}
    \label{tab:specific_lags_var}
    \begin{tabular}{cccc}
        \toprule
        BA (\textuparrow) & $T = 250$ & $T = 500$ & $T = 1000$ \\
        \midrule
        $V = 10$ & 93.1\resSmall{±3.0} & 92.5\resSmall{±1.0} & 95.5\resSmall{±1.0} \\
        $V = 20$ & 83.9\resSmall{±1.0} & 89.1\resSmall{±1.5} & 88.5\resSmall{±2.0} \\
        \bottomrule
    \end{tabular}
    \vspace{-\baselineskip}
\end{wraptable}
\densePar{Variable Number of Lags.}
\name{} can naturally be extended to learn separate projections $\bm{W}^{(\ell)}$ per time lag $\ell$, effectively inflating $\bm{W}$ to rank three.
We evaluate it on the simulated VAR dataset across different time series lengths $T$ and number of variates $V$ in \Cref{tab:specific_lags_var}, proving its ability to learn multiple lag-specific relations without training additional xLSTM models.

\denseParActually{Complexity and Scaling Behaviour.}
To discover the entire GC graph, we need to fit $V$ models $\mathcal{M}_{\bm{\theta}_v,v}$ to obtain the respective sparse projection matrices $\bm{W}_v$ containing all edges arriving at each $v$.
Assuming for brevity that the latent dimension $D \propto V$, i.e., is a fixed multiple of the number of variates $V$, each fitting runs in $O(TV^2/N_h)$ time and requires $O(TV^2/N_h)$ space, with $T$ being the time steps and $N_h$ the number of sLSTM heads.
Their block-wise structure permits that, despite the squared effort, forward and backward passes are efficiently computable even for thousands of dimensions.
Depending on whether all $V$ models are estimated in sequence or parallel, either the time or memory complexity multiplies by $V$ to arrive at the total cost.
In practice, however, \name{} is extremely efficient on contemporary computing platforms due to the availability of highly optimized implementations for xLSTM layers.
\Cref{fig:scaling_exp} in \Cref{sec:app:scaling} shows that it effectively scales linearly in the number of variates for the ranges relevant to standard GC detection settings.

\denseParActually{Inspecting Training Dynamics.}
Here, we elaborate on the utility of the logarithm in the reduction loss $\mathcal{L}_\text{red}$ of \Cref{eq:loss_gd}.
The logarithm term incentivizes the model to explore sparser solutions to GC discovery by allowing the training to move forward over any local minima that use the complete set of input variates.
As the projection matrix becomes sparser and the input variates vanish from consideration by the model, the prediction loss $\mathcal{L}_\text{pred}$ will slightly increase. 
As seen in \Cref{fig:training_combined} in \Cref{sec:app:training_dynamics}, an increase in the sparsity of the feature selectors $\bm W$ drives down the loss and enables learning more meaningful GC relations. It also shows how the variable usage quickly stabilizes.

\section{Conclusion}
\label{sec:conclusion}

We presented \name{}, a novel xLSTM-based model to uncover Granger causal relations from the underlying time series data. \name{} first enforces sparsity between the time series components and then learns a weight per time series to decide the importance of each time series for the underlying task. Each time series component is then modeled using a separate xLSTM model, which enables it to better discover Granger causal relationships between the time series variables. We validated \name{} in six scenarios, showing its effectiveness and adaptability in uncovering Granger causal relationships even in the presence of complex and noisy data.

\denseParActually{Limitations.}
While \Cref{sec:method:theoretical_analysis} provides a theoretical analysis of \name{}, it does not give guarantees in the form of mathematical proofs.
The rigor is limited by a lack of formal analysis of xLSTM blocks (see also \Cref{sec:app:xlstm_approx}), which are yet to be formally analyzed to the same degree as more established architectures like LSTMs.
While we discuss and measure the scaling behaviour of \name{} in \Cref{sec:exp:model_analysis}, we only focused on dozens of variates as common in the literature.

\denseParActually{Future work.} This includes using more sophisticated architectures such as TimeMixer~\citep{wangTimeMixerDecomposableMultiscale2024} or xLSTM-Mixer~\citep{krausXLSTMMixerMultivariateTime2025}.
Furthermore, discovering causal links specific to certain lags could be refined, where per-lag projections are learned for the near past and a remainder projection for the more distant lags.
Finally, extending our evaluations to more real-world datasets encompassing domains such as climate change or ecology is an essential next step.

%% file: content/checklist.tex
\section*{NeurIPS Paper Checklist}

\begin{enumerate}

\item {\bf Claims}
    \item[] Question: Do the main claims made in the abstract and introduction accurately reflect the paper's contributions and scope?
    \item[] Answer: \answerYes{}
    \item[] Justification: We explicitly list the contributions at the end of \Cref{sec:intro}.
    \item[] Guidelines:
    \begin{itemize}
        \item The answer NA means that the abstract and introduction do not include the claims made in the paper.
        \item The abstract and/or introduction should clearly state the claims made, including the contributions made in the paper and important assumptions and limitations. A No or NA answer to this question will not be perceived well by the reviewers. 
        \item The claims made should match theoretical and experimental results, and reflect how much the results can be expected to generalize to other settings. 
        \item It is fine to include aspirational goals as motivation as long as it is clear that these goals are not attained by the paper. 
    \end{itemize}

\item {\bf Limitations}
    \item[] Question: Does the paper discuss the limitations of the work performed by the authors?
    \item[] Answer: \answerYes{}
    \item[] Justification: We discuss limitations as an explicitly named paragraph in \Cref{sec:conclusion}. We acknowledge that any empirical results are by their very nature limited to the settings in which they were obtained, and thus strive to accurately describe them for best reproducibility (see also Question 4).
    \item[] Guidelines:
    \begin{itemize}
        \item The answer NA means that the paper has no limitation while the answer No means that the paper has limitations, but those are not discussed in the paper. 
        \item The authors are encouraged to create a separate "Limitations" section in their paper.
        \item The paper should point out any strong assumptions and how robust the results are to violations of these assumptions (e.g., independence assumptions, noiseless settings, model well-specification, asymptotic approximations only holding locally). The authors should reflect on how these assumptions might be violated in practice and what the implications would be.
        \item The authors should reflect on the scope of the claims made, e.g., if the approach was only tested on a few datasets or with a few runs. In general, empirical results often depend on implicit assumptions, which should be articulated.
        \item The authors should reflect on the factors that influence the performance of the approach. For example, a facial recognition algorithm may perform poorly when image resolution is low or images are taken in low lighting. Or a speech-to-text system might not be used reliably to provide closed captions for online lectures because it fails to handle technical jargon.
        \item The authors should discuss the computational efficiency of the proposed algorithms and how they scale with dataset size.
        \item If applicable, the authors should discuss possible limitations of their approach to address problems of privacy and fairness.
        \item While the authors might fear that complete honesty about limitations might be used by reviewers as grounds for rejection, a worse outcome might be that reviewers discover limitations that aren't acknowledged in the paper. The authors should use their best judgment and recognize that individual actions in favor of transparency play an important role in developing norms that preserve the integrity of the community. Reviewers will be specifically instructed to not penalize honesty concerning limitations.
    \end{itemize}

\item {\bf Theory assumptions and proofs}
    \item[] Question: For each theoretical result, does the paper provide the full set of assumptions and a complete (and correct) proof?
    \item[] Answer: \answerNA{}
    \item[] Justification: While we perform theoretical anlaysis in \Cref{sec:method:theoretical_analysis,sec:app:xlstm_approx}, these are not in the form of mathematical theorems.
    \item[] Guidelines:
    \begin{itemize}
        \item The answer NA means that the paper does not include theoretical results. 
        \item All the theorems, formulas, and proofs in the paper should be numbered and cross-referenced.
        \item All assumptions should be clearly stated or referenced in the statement of any theorems.
        \item The proofs can either appear in the main paper or the supplemental material, but if they appear in the supplemental material, the authors are encouraged to provide a short proof sketch to provide intuition. 
        \item Inversely, any informal proof provided in the core of the paper should be complemented by formal proofs provided in appendix or supplemental material.
        \item Theorems and Lemmas that the proof relies upon should be properly referenced. 
    \end{itemize}

    \item {\bf Experimental result reproducibility}
    \item[] Question: Does the paper fully disclose all the information needed to reproduce the main experimental results of the paper to the extent that it affects the main claims and/or conclusions of the paper (regardless of whether the code and data are provided or not)?
    \item[] Answer: \answerYes{}
    \item[] Justification: All implementation details, including dataset descriptions and experiment configurations, are provided in \Cref{sec:exp}.
    \item[] Guidelines:
    \begin{itemize}
        \item The answer NA means that the paper does not include experiments.
        \item If the paper includes experiments, a No answer to this question will not be perceived well by the reviewers: Making the paper reproducible is important, regardless of whether the code and data are provided or not.
        \item If the contribution is a dataset and/or model, the authors should describe the steps taken to make their results reproducible or verifiable. 
        \item Depending on the contribution, reproducibility can be accomplished in various ways. For example, if the contribution is a novel architecture, describing the architecture fully might suffice, or if the contribution is a specific model and empirical evaluation, it may be necessary to either make it possible for others to replicate the model with the same dataset, or provide access to the model. In general. releasing code and data is often one good way to accomplish this, but reproducibility can also be provided via detailed instructions for how to replicate the results, access to a hosted model (e.g., in the case of a large language model), releasing of a model checkpoint, or other means that are appropriate to the research performed.
        \item While NeurIPS does not require releasing code, the conference does require all submissions to provide some reasonable avenue for reproducibility, which may depend on the nature of the contribution. For example
        \begin{enumerate}
            \item If the contribution is primarily a new algorithm, the paper should make it clear how to reproduce that algorithm.
            \item If the contribution is primarily a new model architecture, the paper should describe the architecture clearly and fully.
            \item If the contribution is a new model (e.g., a large language model), then there should either be a way to access this model for reproducing the results or a way to reproduce the model (e.g., with an open-source dataset or instructions for how to construct the dataset).
            \item We recognize that reproducibility may be tricky in some cases, in which case authors are welcome to describe the particular way they provide for reproducibility. In the case of closed-source models, it may be that access to the model is limited in some way (e.g., to registered users), but it should be possible for other researchers to have some path to reproducing or verifying the results.
        \end{enumerate}
    \end{itemize}

\item {\bf Open access to data and code}
    \item[] Question: Does the paper provide open access to the data and code, with sufficient instructions to faithfully reproduce the main experimental results, as described in supplemental material?
    \item[] Answer: \answerYes{}
    \item[] Justification: We strive to use freely available datasets where possible and exclusively employ openly available software. Furthermore, we provide the source code for full reproducibility at \codeLink{}.
    \item[] Guidelines:
    \begin{itemize}
        \item The answer NA means that paper does not include experiments requiring code.
        \item Please see the NeurIPS code and data submission guidelines (\url{https://nips.cc/public/guides/CodeSubmissionPolicy}) for more details.
        \item While we encourage the release of code and data, we understand that this might not be possible, so “No” is an acceptable answer. Papers cannot be rejected simply for not including code, unless this is central to the contribution (e.g., for a new open-source benchmark).
        \item The instructions should contain the exact command and environment needed to run to reproduce the results. See the NeurIPS code and data submission guidelines (\url{https://nips.cc/public/guides/CodeSubmissionPolicy}) for more details.
        \item The authors should provide instructions on data access and preparation, including how to access the raw data, preprocessed data, intermediate data, and generated data, etc.
        \item The authors should provide scripts to reproduce all experimental results for the new proposed method and baselines. If only a subset of experiments are reproducible, they should state which ones are omitted from the script and why.
        \item At submission time, to preserve anonymity, the authors should release anonymized versions (if applicable).
        \item Providing as much information as possible in supplemental material (appended to the paper) is recommended, but including URLs to data and code is permitted.
    \end{itemize}

\item {\bf Experimental setting/details}
    \item[] Question: Does the paper specify all the training and test details (e.g., data splits, hyperparameters, how they were chosen, type of optimizer, etc.) necessary to understand the results?
    \item[] Answer: \answerYes{}
    \item[] Justification: All such details are provided in \Cref{sec:exp}.
    \item[] Guidelines:
    \begin{itemize}
        \item The answer NA means that the paper does not include experiments.
        \item The experimental setting should be presented in the core of the paper to a level of detail that is necessary to appreciate the results and make sense of them.
        \item The full details can be provided either with the code, in appendix, or as supplemental material.
    \end{itemize}

\item {\bf Experiment statistical significance}
    \item[] Question: Does the paper report error bars suitably and correctly defined or other appropriate information about the statistical significance of the experiments?
    \item[] Answer: \answerYes{}
    \item[] Justification: All quantitative results are accompanied by standard deviations (specifically, see \Cref{tab:auroc_lorenz}).
    \item[] Guidelines:
    \begin{itemize}
        \item The answer NA means that the paper does not include experiments.
        \item The authors should answer "Yes" if the results are accompanied by error bars, confidence intervals, or statistical significance tests, at least for the experiments that support the main claims of the paper.
        \item The factors of variability that the error bars are capturing should be clearly stated (for example, train/test split, initialization, random drawing of some parameter, or overall run with given experimental conditions).
        \item The method for calculating the error bars should be explained (closed form formula, call to a library function, bootstrap, etc.)
        \item The assumptions made should be given (e.g., Normally distributed errors).
        \item It should be clear whether the error bar is the standard deviation or the standard error of the mean.
        \item It is OK to report 1-sigma error bars, but one should state it. The authors should preferably report a 2-sigma error bar than state that they have a 96\% CI, if the hypothesis of Normality of errors is not verified.
        \item For asymmetric distributions, the authors should be careful not to show in tables or figures symmetric error bars that would yield results that are out of range (e.g. negative error rates).
        \item If error bars are reported in tables or plots, The authors should explain in the text how they were calculated and reference the corresponding figures or tables in the text.
    \end{itemize}

\item {\bf Experiments compute resources}
    \item[] Question: For each experiment, does the paper provide sufficient information on the computer resources (type of compute workers, memory, time of execution) needed to reproduce the experiments?
    \item[] Answer: \answerYes{}
    \item[] Justification: This is provided at the beginning of \Cref{sec:exp}.
    \item[] Guidelines:
    \begin{itemize}
        \item The answer NA means that the paper does not include experiments.
        \item The paper should indicate the type of compute workers CPU or GPU, internal cluster, or cloud provider, including relevant memory and storage.
        \item The paper should provide the amount of compute required for each of the individual experimental runs as well as estimate the total compute. 
        \item The paper should disclose whether the full research project required more compute than the experiments reported in the paper (e.g., preliminary or failed experiments that didn't make it into the paper). 
    \end{itemize}
    
\item {\bf Code of ethics}
    \item[] Question: Does the research conducted in the paper conform, in every respect, with the NeurIPS Code of Ethics \url{https://neurips.cc/public/EthicsGuidelines}?
    \item[] Answer: \answerYes{}
    \item[] Justification: We made sure to comply with the Code of Ethics.
    \item[] Guidelines:
    \begin{itemize}
        \item The answer NA means that the authors have not reviewed the NeurIPS Code of Ethics.
        \item If the authors answer No, they should explain the special circumstances that require a deviation from the Code of Ethics.
        \item The authors should make sure to preserve anonymity (e.g., if there is a special consideration due to laws or regulations in their jurisdiction).
    \end{itemize}

\item {\bf Broader impacts}
    \item[] Question: Does the paper discuss both potential positive societal impacts and negative societal impacts of the work performed?
    \item[] Answer: \answerNA{}
    \item[] Justification: While we acknowledge that many technologies have wide-ranging societal impacts, our primary focus is on technical innovation. We have thus not identified specific concerns requiring emphasis in this work.
    \item[] Guidelines:
    \begin{itemize}
        \item The answer NA means that there is no societal impact of the work performed.
        \item If the authors answer NA or No, they should explain why their work has no societal impact or why the paper does not address societal impact.
        \item Examples of negative societal impacts include potential malicious or unintended uses (e.g., disinformation, generating fake profiles, surveillance), fairness considerations (e.g., deployment of technologies that could make decisions that unfairly impact specific groups), privacy considerations, and security considerations.
        \item The conference expects that many papers will be foundational research and not tied to particular applications, let alone deployments. However, if there is a direct path to any negative applications, the authors should point it out. For example, it is legitimate to point out that an improvement in the quality of generative models could be used to generate deepfakes for disinformation. On the other hand, it is not needed to point out that a generic algorithm for optimizing neural networks could enable people to train models that generate Deepfakes faster.
        \item The authors should consider possible harms that could arise when the technology is being used as intended and functioning correctly, harms that could arise when the technology is being used as intended but gives incorrect results, and harms following from (intentional or unintentional) misuse of the technology.
        \item If there are negative societal impacts, the authors could also discuss possible mitigation strategies (e.g., gated release of models, providing defenses in addition to attacks, mechanisms for monitoring misuse, mechanisms to monitor how a system learns from feedback over time, improving the efficiency and accessibility of ML).
    \end{itemize}
    
\item {\bf Safeguards}
    \item[] Question: Does the paper describe safeguards that have been put in place for responsible release of data or models that have a high risk for misuse (e.g., pretrained language models, image generators, or scraped datasets)?
    \item[] Answer: \answerNA{}
    \item[] Justification: Please see Question 10. Specifically, we do not provide any trained models or similar high-risk artifacts.
    \item[] Guidelines:
    \begin{itemize}
        \item The answer NA means that the paper poses no such risks.
        \item Released models that have a high risk for misuse or dual-use should be released with necessary safeguards to allow for controlled use of the model, for example by requiring that users adhere to usage guidelines or restrictions to access the model or implementing safety filters. 
        \item Datasets that have been scraped from the Internet could pose safety risks. The authors should describe how they avoided releasing unsafe images.
        \item We recognize that providing effective safeguards is challenging, and many papers do not require this, but we encourage authors to take this into account and make a best faith effort.
    \end{itemize}

\item {\bf Licenses for existing assets}
    \item[] Question: Are the creators or original owners of assets (e.g., code, data, models), used in the paper, properly credited and are the license and terms of use explicitly mentioned and properly respected?
    \item[] Answer: \answerYes{}
    \item[] Justification: We carefully cite all immediately relevant scholarly works and provide URLs to any other resources in \Cref{sec:exp,sec:app:datasets}.
    \item[] Guidelines:
    \begin{itemize}
        \item The answer NA means that the paper does not use existing assets.
        \item The authors should cite the original paper that produced the code package or dataset.
        \item The authors should state which version of the asset is used and, if possible, include a URL.
        \item The name of the license (e.g., CC-BY 4.0) should be included for each asset.
        \item For scraped data from a particular source (e.g., website), the copyright and terms of service of that source should be provided.
        \item If assets are released, the license, copyright information, and terms of use in the package should be provided. For popular datasets, \url{paperswithcode.com/datasets} has curated licenses for some datasets. Their licensing guide can help determine the license of a dataset.
        \item For existing datasets that are re-packaged, both the original license and the license of the derived asset (if it has changed) should be provided.
        \item If this information is not available online, the authors are encouraged to reach out to the asset's creators.
    \end{itemize}

\item {\bf New assets}
    \item[] Question: Are new assets introduced in the paper well documented and is the documentation provided alongside the assets?
    \item[] Answer: \answerYes{}
    \item[] Justification: The documentation of the source code is also provided at \codeLink{}.
    \item[] Guidelines:
    \begin{itemize}
        \item The answer NA means that the paper does not release new assets.
        \item Researchers should communicate the details of the dataset/code/model as part of their submissions via structured templates. This includes details about training, license, limitations, etc. 
        \item The paper should discuss whether and how consent was obtained from people whose asset is used.
        \item At submission time, remember to anonymize your assets (if applicable). You can either create an anonymized URL or include an anonymized zip file.
    \end{itemize}

\item {\bf Crowdsourcing and research with human subjects}
    \item[] Question: For crowdsourcing experiments and research with human subjects, does the paper include the full text of instructions given to participants and screenshots, if applicable, as well as details about compensation (if any)? 
    \item[] Answer: \answerNA{}
    \item[] Justification: We did not perform any such experiments.
    \item[] Guidelines:
    \begin{itemize}
        \item The answer NA means that the paper does not involve crowdsourcing nor research with human subjects.
        \item Including this information in the supplemental material is fine, but if the main contribution of the paper involves human subjects, then as much detail as possible should be included in the main paper. 
        \item According to the NeurIPS Code of Ethics, workers involved in data collection, curation, or other labor should be paid at least the minimum wage in the country of the data collector. 
    \end{itemize}

\item {\bf Institutional review board (IRB) approvals or equivalent for research with human subjects}
    \item[] Question: Does the paper describe potential risks incurred by study participants, whether such risks were disclosed to the subjects, and whether Institutional Review Board (IRB) approvals (or an equivalent approval/review based on the requirements of your country or institution) were obtained?
    \item[] Answer: \answerNA{}
    \item[] Justification: Please see Question 14.
    \item[] Guidelines:
    \begin{itemize}
        \item The answer NA means that the paper does not involve crowdsourcing nor research with human subjects.
        \item Depending on the country in which research is conducted, IRB approval (or equivalent) may be required for any human subjects research. If you obtained IRB approval, you should clearly state this in the paper. 
        \item We recognize that the procedures for this may vary significantly between institutions and locations, and we expect authors to adhere to the NeurIPS Code of Ethics and the guidelines for their institution. 
        \item For initial submissions, do not include any information that would break anonymity (if applicable), such as the institution conducting the review.
    \end{itemize}

\item {\bf Declaration of LLM usage}
    \item[] Question: Does the paper describe the usage of LLMs if it is an important, original, or non-standard component of the core methods in this research? Note that if the LLM is used only for writing, editing, or formatting purposes and does not impact the core methodology, scientific rigorousness, or originality of the research, declaration is not required.
    \item[] Answer: \answerNA{}
    \item[] Justification: We did not employ LLMs in any part of \name{}.
    \item[] Guidelines:
    \begin{itemize}
        \item The answer NA means that the core method development in this research does not involve LLMs as any important, original, or non-standard components.
        \item Please refer to our LLM policy (\url{https://neurips.cc/Conferences/2025/LLM}) for what should or should not be described.
    \end{itemize}

\end{enumerate}

%% file: content/appendix.tex
\section{Full Algorithm}

\Cref{alg:main} describes the entire procedure for extracting a Granger causal graph using \name{}.
$(\cdot)_+$ denotes truncation as $\max(\cdot, 0)$.

\begin{algorithm}[h]
    \caption{Granger causality detection with \name{}}
    \label{alg:main}
    \begin{algorithmic}[1]
        \Input{
            Training data $\mathcal{D} = \left\{ \bm{S}^{(i)} \right\}_{i \in \{1,\dots,N \}}$,
            sparsity hyperparameter $\lambda \in \R_+$,
            learning rate $\eta \in \R_+$, and
            compression schedule start $K \in \N$
        }
        \Output{Granger causal graph}
        \algorithmicbreak
        \ForAll{$v \in \mathcal{V}$} \Comment{Training decomposes over $\mathcal V$}
            \State{$\bm{\phi}_v : \bm{W}_v, \bm{b}_v \sim \mathcal{U}\left(-\sqrt{\nicefrac{1}{V}}, \sqrt{\nicefrac{1}{V}}\right)$} \Comment{Kaiming/He initialization}
            \State{$\bm{\alpha}_v \gets \nicefrac{1}{V} \mathds{1}$} \Comment{Uniform reduction coefficient}
            \State{$\bm{\theta}_v \gets \bm{\theta}^{(0)}$} \Comment{Standard xLSTM initialization}
            \State{$k \gets 0$}
            \Repeat
                \State{Sample random mini-batch $\mathcal{B} \sim \mathcal{D}$}
                \State{$\bm{g}_{\bm\phi, \bm\alpha, \bm\theta} \gets \nabla_{\bm{\phi}, \bm{\alpha}, \bm{\theta}} \frac{1}{|\mathcal{B}|} \sum_{\bm S \in \mathcal{B}} \Big[ \mathcal{L}_\text{pred}(\bm S ; \bm{\phi}_v, \bm{\theta}_v) + \lambda \log\left(\sum_{w=1}^V \alpha_v^w \norm{\sg\left( \bm{W}_v^w \right)}_2 \right) \Big]$} \label{alg:main:gradient_start} %
                \State{$\bm{\phi}_v \gets \bm{\phi}_v - \eta \bm{g}_{\bm\phi}$} \Comment{GD step}
                \State{$\bm{\theta}_v \gets \bm{\theta}_v - \eta \bm{g}_{\bm\theta}$} \Comment{GD step}
                \If{$k \geq K$} \Comment{Optimize $\bm\alpha$ after $K$ steps}  \label{alg:main:staged_alpha}
                    \State{$\bm{\alpha}_v \gets \bm{\alpha}_v - \eta \bm{g}_{\bm\alpha}$} \Comment{GD step}
                \EndIf \label{alg:main:gradient_end}
                \State{$\bm{W}_v \gets \left( 1 - \frac{\lambda \eta \bm{\alpha}_v}{\norm{\bm{W}_v}_2} \right)_+ \bm{W}_v$} \Comment{Compression step with proximal GD} \label{alg:main:prox}
                \State{$k \gets k + 1$}
            \Until{convergence}
        \EndFor
        \State{$\mathcal E \gets \left\{ (v, w) \in \mathcal{V} \times \mathcal{V} \; \middle| \; \norm{\bm{W}_v^w}_2 > 0 \right\}$}
        \State\Return{extracted graph $(\mathcal V, \mathcal E)$}
    \end{algorithmic}
\end{algorithm}

\section{On the Approximation Capabilities of sLSTM blocks}
\label{sec:app:xlstm_approx}

Classic RNNs have long been known to be extremely powerful models of computation. Specifically, they are Turing complete~\citep{siegelmannComputationalPowerNeural1995} and, by extension, universal function approximators~\citep{songMinimalWidthUniversal2023}.
Traditional LSTMs as proposed by \citet{hochreiterLongShortTermMemory1997} are universal function approximators being at least as powerful as the RNNs \citep[Corollary~16]{songMinimalWidthUniversal2023}.
Due to their novelty, the sLSTM \emph{cells} as presented in \Cref{sec:prelim_related:xlstm} have not yet received the same degree of theoretical analysis.\footnote{A discussion of expressivity hierarchies can be found in \citet[App.~A]{auerTiRexZeroShotForecasting2025}.}
Yet, it appears natural to extend the same reduction to RNNs as has been shown for LSTMs, since the main technicality is the differing normalization of the hidden state~\citep[Eq.~36]{beckXLSTMExtendedLong2024} and the alternative exponential activation function, which \citet{songMinimalWidthUniversal2023} abstract away with.
sLSTM \emph{blocks} then wrap these cells with additional operations~(cf. \citet{beckXLSTMExtendedLong2024}, App.~A.4).
They, however, can be carefully configured to reduce to only rescaling Layer Normalizations and residual connections, which are known to, again, yield universal function approximators for many architectures~\citep{tabuadaUniversalApproximationPower2020,marchiTrainingDeepResidual2021}.
This can be shown by first omitting the optional upfront convolution and Swish activation.
We, furthermore, set the number of heads to $N_h = 1$, causing the per-head Group Normalization after the cell to degenerate to yet another Layer Normalization.
Lastly, the final post up-projection of $\bm{x}_\text{in}$ to $\bm{x}_\text{out}$ is defined as
$$
    \bm{x}_\text{out} = \bm{W}_3 \left( \left(\bm{W}_1 \bm{x}_\text{in} + \bm{b}_1 \right) \odot \GeLU\left(\bm{W}_2 \bm{x}_\text{in} + \bm{b}_2 \right) \right) + \bm{b}_3
    .
$$
With the right instantiation, where $\alpha$ is chosen such that $\GeLU(\alpha) = 1$, we obtain
\begin{align*}
    \bm{x}_\text{out}
    &= 
    \begin{pmatrix}
        \I & 0  & 0  & 0 \\
        0  & \I & 0  & 0 \\
        0  & 0  & \I & 0
    \end{pmatrix}
    \left(
        \left(\begin{pmatrix}
            \I & 0  & 0 \\
            0  & \I & 0 \\
            0  & 0  & \I \\
            0  & 0  & 0
        \end{pmatrix} \bm{x}_\text{in} + \bm{0} \right)
        \odot
        \GeLU\left( \bm{0} \bm{x}_\text{in} + \alpha\1 \right)
    \right)
    + \bm{0} \\
    &= 
    \begin{pmatrix}
        \I & 0  & 0  & 0 \\
        0  & \I & 0  & 0 \\
        0  & 0  & \I & 0
    \end{pmatrix}
    \begin{pmatrix}
        \I & 0  & 0 \\
        0  & \I & 0 \\
        0  & 0  & \I \\
        0  & 0  & 0
    \end{pmatrix} \bm{x}_\text{in}
    \odot \1 \\
    &= \bm{x}_\text{in} .
\end{align*}
This, again, shows that sLSTM blocks are at least as general as LSTMs.
While a rigorous proof is far beyond the scope of this work, this investigation still underpins the strong capabilities of this architecture.
These theoretical considerations align well with empirical findings, showing that xLSTM blocks are at least as effective as LSTMs \citep{beckXLSTMExtendedLong2024}.

\section{Dataset Details}
\label{sec:app:datasets}

This section details all six datasets used in our empirical evaluation of \name{}. An overview is provided in \Cref{tab:dataset_details}.

\begin{table}[ht]
    \centering
    \caption{\textbf{Overview of the diverse collection of datasets.} It also shows the number of variates $V$, time steps $T$, samples $N$, and look-back context steps of \name{} $C$.}
    \label{tab:dataset_details}
    \tableSkip
    \begin{tabular}{lllrrrr}
        \toprule
        Name & Origin & Type & $V$ & $T$ & $N$ & $C$ \\
        \midrule
        Lorenz-96 & \citet{Karimi_2010} & Simulated & 20 & 500 & 1 & 10 \\
        fMRI & \citet{smithNetworkModellingMethods2011} & Real-world & 15 & 200 & 1 & 10 \\
        Moléne & \citet{molene2015dataset} & Real-world & 32 & 744 & 1 & 10 \\
        Human MoCap (Run) & \citet{cmu_mocap} & Real-world & 54 & 1232 & 61 & 10 \\
        Human MoCap (Salsa) & \citet{cmu_mocap} & Real-world & 54 & 4136 & 30 & 10 \\
        Company Fundamentals & \citet{divoForecastingCompanyFundamentals2025} & Real-world & 19 & 56 & 2527 & 40 \\
        VAR & \citet{Karimi_2010} & Simulated & 10/20 & mult. & 1 & 5 \\
        \bottomrule
    \end{tabular}
\end{table}

\paragraph{Lorenz-96.}
The $V$-dimensional Lorenz-96 model~\citep{Karimi_2010} is a chaotic multivariate dynamical system governed by the differential equations 
$$
    \frac{d x_{t, i}}{dt} = (x_{t, i+1} - x_{t, i-2})x_{t, i-1} - x_{t, i} + F ,
$$
with the external forcing coefficient $F$ regulating the non-linearity of the system.
Low values of $F$ correspond to near-linear dynamics, while higher values induce chaotic behavior.
There are two sources of randomness in the system. Firstly, we sample i.i.d. starting conditions from $\mathcal{N}(0, 0.01)$.
Secondly, in each step of the simulation, we add i.i.d. noise sampled from $\mathcal{N}(0, 0.1)$ to $x_{t,i}$.
Following the setting of \cite{tank2022neural} for best comparability, we simulate $V=20$ variates with a sampling rate of $\Delta t = 0.05$ for a total of $T = 500$ time steps after a brief burn-in time.
We use two forcing constants $F \in \{ 10, 40 \}$ to test our model under different levels of non-linearity.

\paragraph{fMRI.}
Discovering connectivity networks within (human) brains is a key application of GC detection methods.
To this end, brain activity is measured non-invasively using functional magnetic resonance imaging~(fMRI) over time and grouped into regions between which connections are looked for.
Specifically, we used the realistic simulations of blood-oxygen-level-dependent~(BOLD) deconvolved data of \citet{smithNetworkModellingMethods2011}.

\paragraph{Moléne.}
The Moléne dataset~\citep{molene2015dataset} contains hourly temperatures recorded by sensors at $V = 32$ locations in Brittany, France, during $T = 744$ hours.
The objective is to understand the spatio-temporal dynamics of the temperature and to assess the extent to which the model can uncover complex relationships in weather by considering only local observations.

\paragraph{Human Motion Capture.}
We also apply our methodology to detect complex, nonlinear dependencies in human motion capture~(MoCap) recordings. In contrast to the Lorenz-96 simulated dataset results, this analysis allows us to visualize and interpret the learned network more easily.
We consider a data set from the CMU MoCap database~\citep{cmu_mocap}.
The data comprises $V=54$ joint angle and body position recordings across multiple subjects.
Since some regions, like the neck, have multiple degrees of freedom in both translation and rotation, we consider the GC relations between two joints based on edges between all movement directions.
The motion ranges from locomotion~(e.g., walking) over physical activities such as gymnastics and dancing to day-to-day social interactions.

\paragraph{Company Fundamentals.}
Another common field of application of GC detection methods is discovering links in economic data, where it is otherwise hard to maintain an overview.
We, therefore, benchmark \name{} on company fundamentals data~\citep{divoForecastingCompanyFundamentals2025}.
The dataset contains $V=19$ economic variables, such as the Net Income and the Total Liabilities of 2527 companies.
The data was collected quarterly from 2009 Q1 to 2023 Q3, resulting in only $T=56$ time steps.

\paragraph{VAR.}
Following the well-known setup of \citet{tank2022neural}, we generate a two-step linear auto-regressive process in $V \in \{10,20\}$ variates.
All variables depend on themselves, and an additional three random dependencies are added as targets to be discovered.
The number of time steps varies in $T \in \{250, 500, 1000\}$, and we start recording the data after a brief burn-in time.

\section{Extended Experimental Results}
\label{sec:app:extended_exp}

\begin{figure}[hp]
    \centering
    \includegraphics[width=0.85\linewidth]{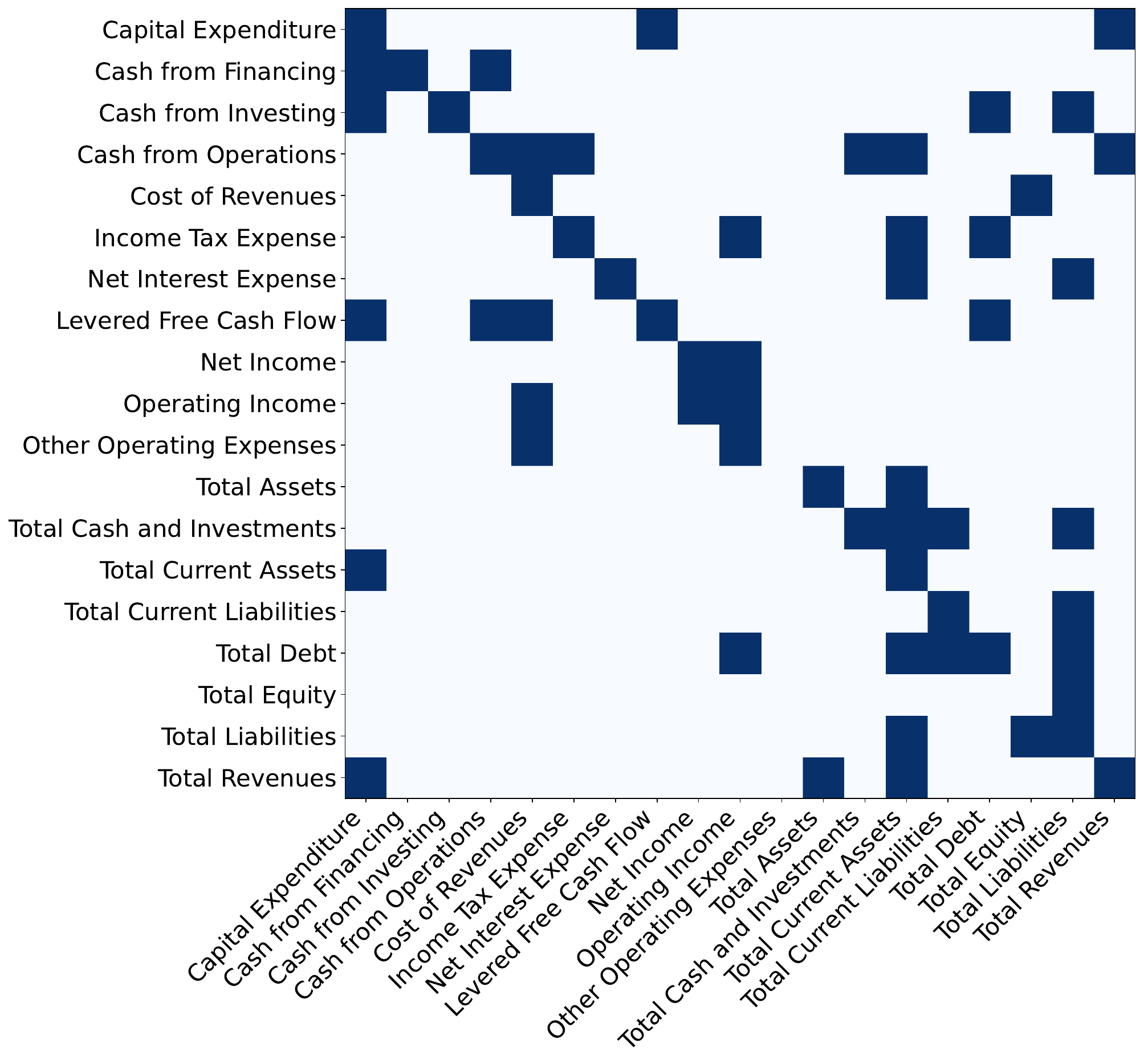}
    \caption{\textbf{\name{} uncovers relationships between economic variables within individual companies.} Established GC links are highlighted in \textcolor[HTML]{132E69}{\textbf{dark blue}}.}
    \label{fig:results_company_fundamentals_matrix}
\end{figure}

This section supplements the experimental findings of \Cref{sec:exp:main}.
Specifically, \Cref{fig:results_company_fundamentals_matrix} provides the complete set of relations extracted from the Company Fundamentals dataset.
On this challenging full feature set, only some of the discovered GC relations are economically plausible.

\section{Scaling Behaviour}
\label{sec:app:scaling}

\begin{figure}[hp]
    \centering
    \includegraphics[width=0.65\linewidth]{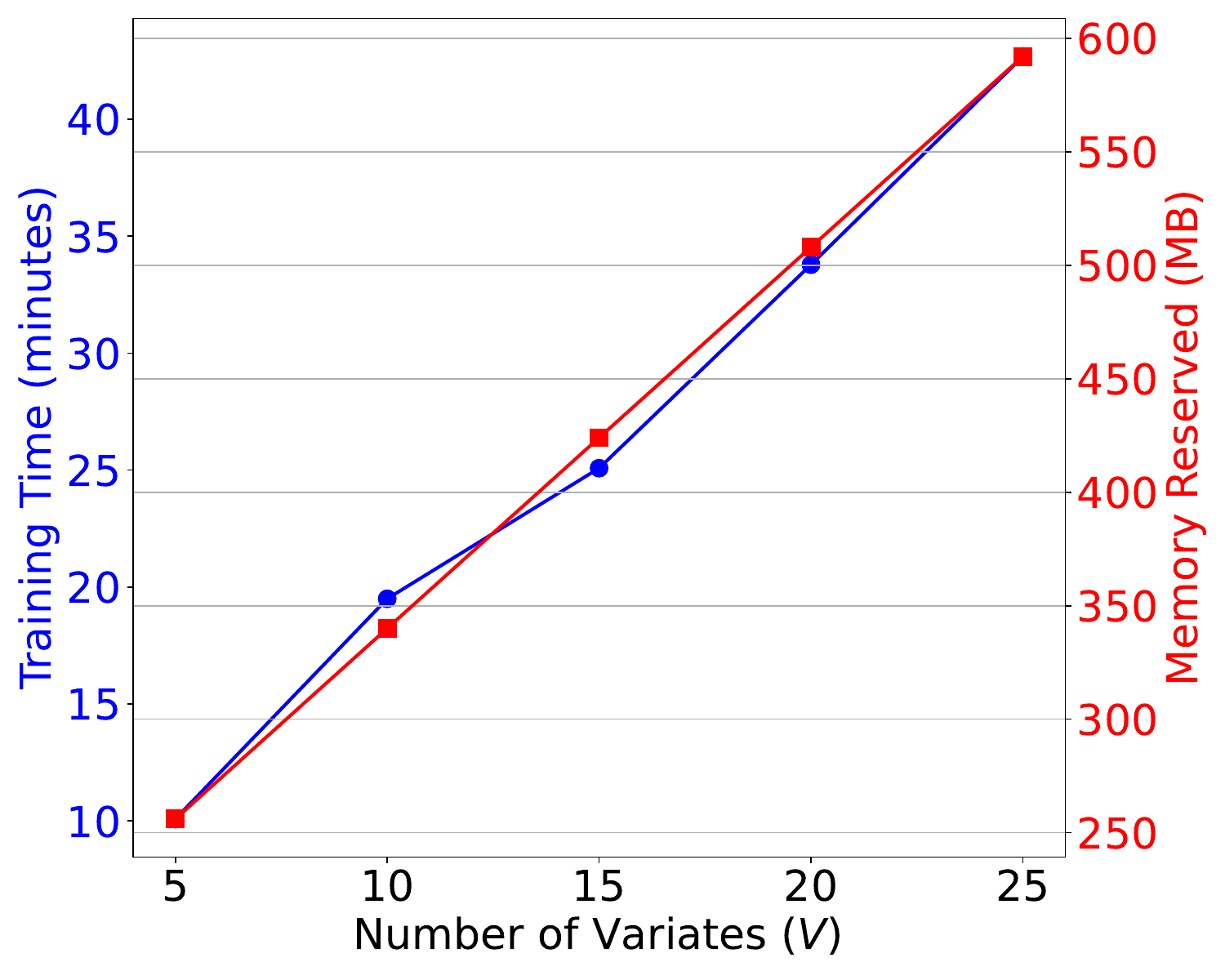}
    \caption{\textbf{\name{} effectively scales linearly in the relevant range of variate counts.}}
    \label{fig:scaling_exp}
\end{figure}

\Cref{fig:scaling_exp} shows the training time and peak GPU memory reserved during the training of \name{} on the Lorenz-96 dataset with $T = 1000$ for various numbers of variates $V$.

\section{Additional Insights into Training}
\label{sec:app:training_dynamics}

\begin{figure}[hp]
    \centering
    \begin{subfigure}[t]{0.46\textwidth}
        \centering
        \includegraphics[width=\textwidth]{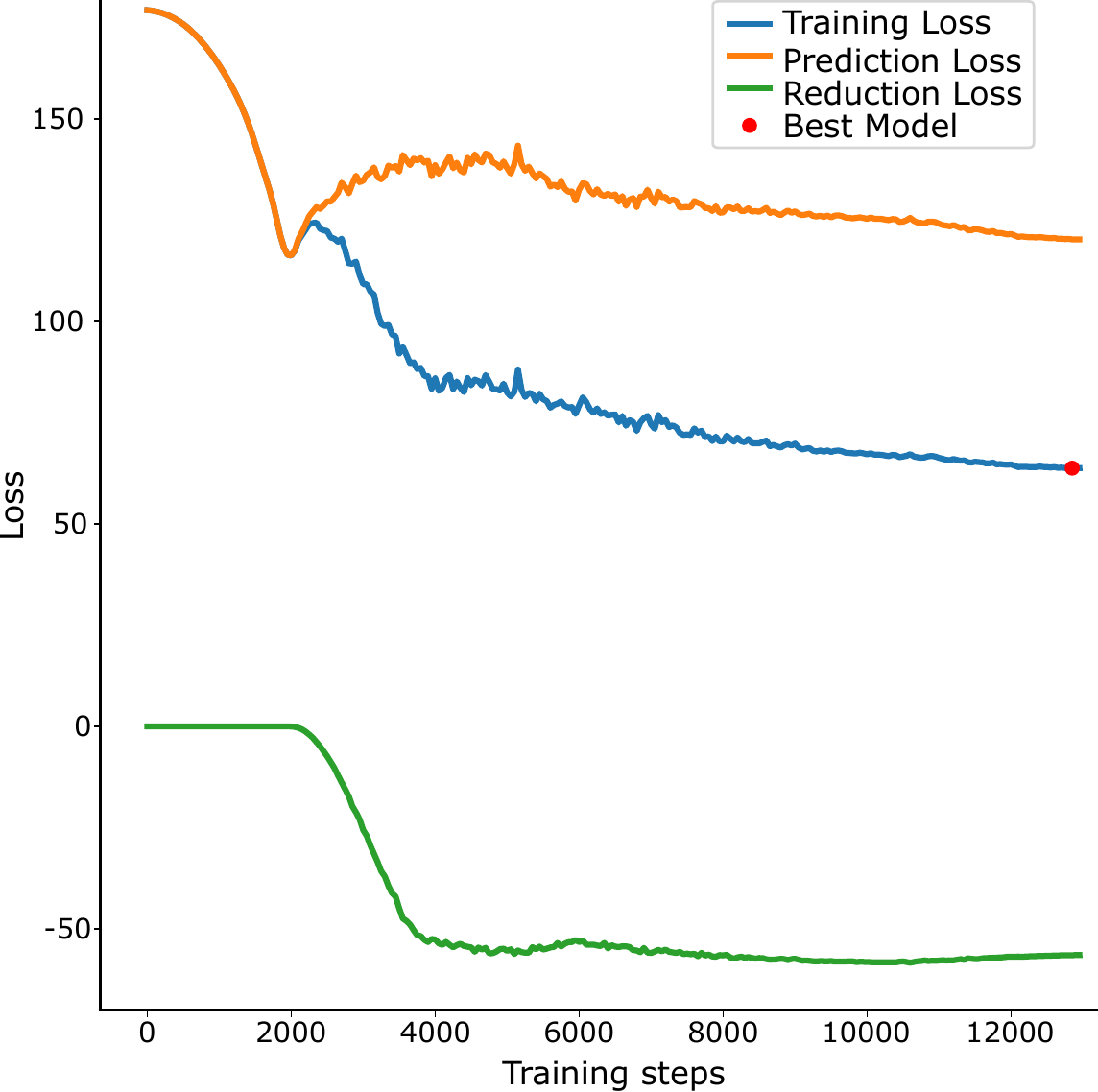}
        \caption{Visualization of the different loss components.}
        \label{fig:losses}
    \end{subfigure}
    \hfill
    \begin{subfigure}[t]{0.50\textwidth}
        \centering
        \includegraphics[width=\textwidth]{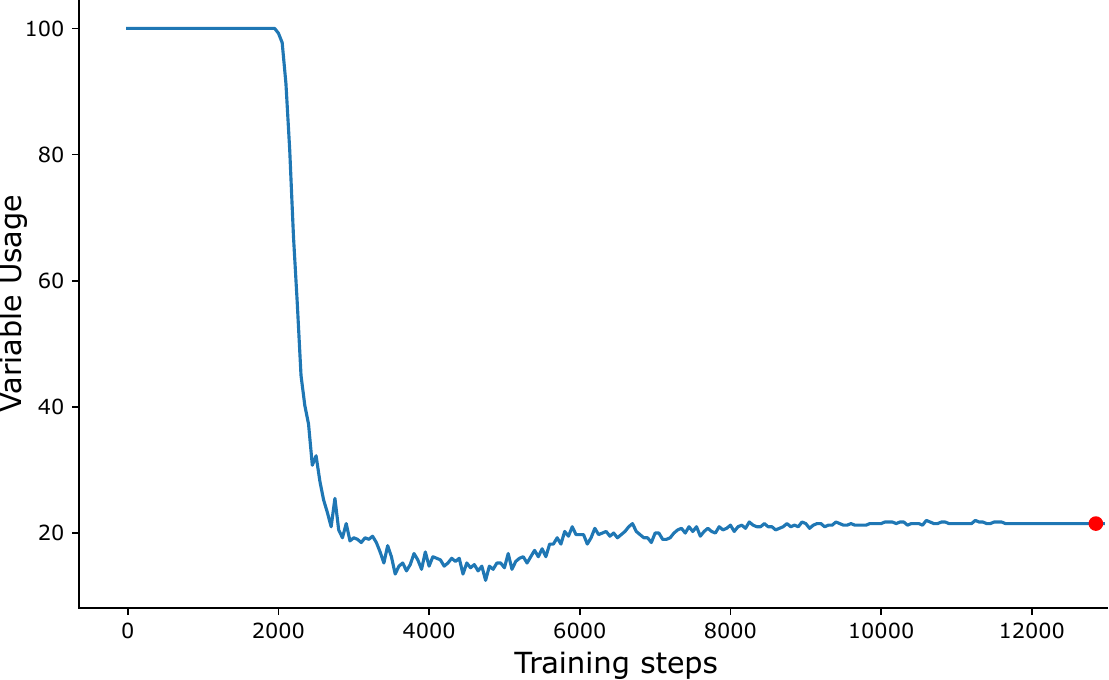}
        \caption{Variable drops once the threshold $k \geq K$ is reached and then quickly stabilizes.}
        \label{fig:variable-usage}
    \end{subfigure}
    \caption{\textbf{\Cref{alg:main} jointly optimizes the prediction loss while adaptively establishing sparsity.} Results show training on Lorenz-96 with $F=40$ and $T=1000$.}
    \label{fig:training_combined}
\end{figure}

\Cref{fig:losses} shows how the different loss components change during training.
The robustness of training is reflected in the stability of variable usage once an optimal set is found, as \Cref{fig:variable-usage} shows.

